\documentclass[10pt,twocolumn,letterpaper]{article}
\ifdefined\pdfsuppressptexinfo \pdfsuppressptexinfo=-1 \fi

\usepackage[pagenumbers]{cvpr}
\usepackage{graphicx}
\usepackage{amsmath,amssymb}
\everymath\expandafter{\the\everymath
  \thickmuskip=3.5mu plus 2mu minus 1mu
  \medmuskip=2.5mu plus 1mu minus 2mu}
\usepackage{booktabs}
\usepackage{xcolor}
\usepackage{colortbl}
\usepackage{multirow}

\AddToHook{env/figure/begin}{\setlength{\belowcaptionskip}{-6pt}}
\AddToHook{env/figure*/begin}{\setlength{\belowcaptionskip}{-6pt}}
\AddToHook{env/subfigure/begin}{\setlength{\belowcaptionskip}{0pt}}
\AddToHook{env/table/end}{\vspace{-6pt}}
\AddToHook{env/table*/end}{\vspace{-6pt}}

\definecolor{cvprblue}{rgb}{0.21,0.49,0.74}
\usepackage[pagebackref,breaklinks,colorlinks,allcolors=cvprblue]{hyperref}

\newcommand{\method}{\textsc{TBD}}   %
\renewcommand{\method}{FFGaussType}

\newcommand{\anchors}{\mathcal{A}}
\newcommand{\plane}{\mathbf{P}}
\newcommand{\enc}{E_{\theta}}
\newcommand{\dec}{D_{\phi}}

\colorlet{bestcolor}{red!25}
\colorlet{secondcolor}{orange!25}
\newcommand{\best}{\cellcolor{bestcolor}}
\newcommand{\second}{\cellcolor{secondcolor}}
\newcommand{\bestkey}[1]{{\setlength{\fboxsep}{1pt}\colorbox{bestcolor}{#1}}}
\newcommand{\secondkey}[1]{{\setlength{\fboxsep}{1pt}\colorbox{secondcolor}{#1}}}

\def\confName{CVPR}
\def\confYear{2027}

\title{\method{}: Feed-Forward Conversion of Glyph and SVG Geometry into Compact Gaussian Assets}
\author{
    Haoan Feng\textsuperscript{1}\thanks{This work was done during the internship at Dolby Laboratories.}, Tianfu Wang\textsuperscript{1}\footnotemark[1], Peimeng Guan\textsuperscript{2}, Dae Yeol Lee\textsuperscript{2}, Guan-ming Su\textsuperscript{2} \\
    \textsuperscript{1}\normalsize{University of Maryland, College Park},\quad \textsuperscript{2}\normalsize{Dolby Laboratories Inc.}\\
    {\tt\small \{hfengac,tianfuw\}@umd.edu, \{pguan, dvlee\}@dolby.com, guanmingsu@ieee.org}\\
    {\small Project page: \href{https://fengyee.github.io/FFGaussType/}{\tt fengyee.github.io/FFGaussType}}
}

\makeatletter
\let\originalmaketitle\@maketitle
\renewcommand{\@maketitle}{%
  \originalmaketitle
  \begin{minipage}{\textwidth}
    \centering
    \includegraphics[width=\linewidth]{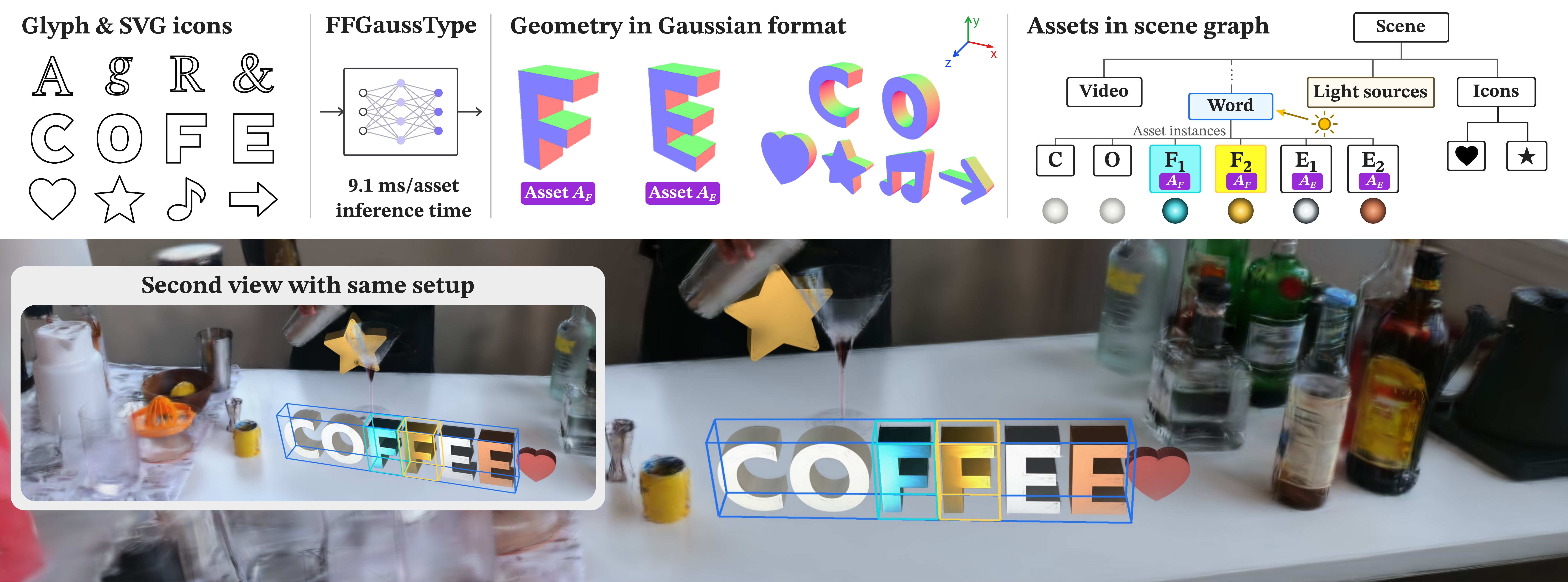}\vspace{0.4em}
    \captionof{figure}{\textbf{Feed-forward conversion and scene-graph reuse of glyph and icon assets.} \method{} converts extruded outlines into Gaussian-format geometry carrying surface normals, shown as normal-colored renderings, in 9.1\,ms per asset. Each converted asset is stored once and instanced in a scene graph, where the letters $\mathrm{F}_1$/$\mathrm{F}_2$ and $\mathrm{E}_1$/$\mathrm{E}_2$ reuse assets $A_F$ and $A_E$ but carry individual materials under a light source shared by the word. The word and icons are composited into the Coffee Martini Gaussian scene~\cite{li2022n3dv}, with bounding boxes colored to match the highlighted instances and a second view of the same setup confirming consistent geometry and shading.}
    \label{fig:teaser}
  \end{minipage}\par\vspace{1em}
}
\makeatother

\begin{document}
\maketitle

\begin{abstract}
Immersive 3D content allows text and graphical elements, such as subtitles, annotations, and icons, to be placed inside a scene rather than displayed as flat overlays.
Placed this way, they respect occlusion and perspective and take on user-assigned materials and lighting while preserving their original shapes.
In Gaussian scenes, however, each new glyph or icon requires its own iterative optimization, which limits scalable content creation.
We present \method{}, which converts extruded glyph and SVG outlines into Gaussian assets in a single forward pass.
A point encoder maps anchors sampled on the known surface to a single feature plane, from which a decoder shared by all assets predicts neural Gaussians carrying surface normals.
This separates geometric conversion from material and lighting choices, and the feature plane supports image and video coding for compact delivery.
We construct a paired dataset of 27,221 glyph and SVG icon assets with anchors and calibrated normal and silhouette renderings, holding out font families for evaluation.
Conversion from prepared anchors takes 9.1\,ms per asset, at a fraction of the compute and memory of prior baselines.
On unseen font families and test icons, \method{} achieves $1.62^\circ$ normal error and $0.985$ silhouette IoU, the best among the compared feed-forward converters, matching the fidelity that per-asset fitting needs thousands of steps to reach.
The code and dataset will be released after publication.
\end{abstract}

\section{Introduction}
\label{sec:intro}

Immersive 3D content lets text and graphical elements, such as subtitles, annotations, and icons, become part of the scene instead of flat overlays on top of it.
Once inserted, they must remain legible as the camera moves and take on the chosen appearance.
Representing them as Gaussian-based assets allows rendering alongside Gaussian-based static scenes~\cite{kerbl20233dgs} and dynamic volumetric videos~\cite{wu20244dgs}, with correct depth occlusion against scene content.
Their shape is already specified by an \emph{outline}, the closed 2D contours of a font glyph or SVG graphic, and its 3D extrusion.
Our task is to convert this known geometry into neural Gaussians that preserve its silhouette and surface normals, leaving material and lighting under the user's control (\cref{fig:teaser}).
We call this task \emph{conversion} because the source geometry is supplied.

The known geometry provides supervision, but the Gaussians still need positions, extents, orientations, and opacities whose blend reproduces the surface.
This is hardest at the \emph{rim}, where the flat front and back faces, or \emph{caps}, meet the narrow side walls and the surface normal changes abruptly.
Per-scene optimization~\cite{kerbl20233dgs} can fit each asset, but it repeats thousands of iterations for every new glyph or icon.
We instead train one converter on many fonts and icon collections and apply it to a new asset in a single forward pass.
Existing feed-forward Gaussian models infer assets from images~\cite{szymanowicz2024splatterimage,tang2024lgm,zhang2024gslrm,zou2024triplanegaussian}, and TexGaussian~\cite{xiong2025texgaussian} predicts appearance on a given surface. Our converter instead takes the known geometry as input and is trained to reproduce its rendered normals and silhouettes.

Our \method{} builds on anchor-based decoding~\cite{lu2024scaffoldgs} and the feature planes of TV-GS~\cite{kathariya2026tv}.
A point encoder maps precomputed surface anchors to a single 2D feature plane defined over the outline, and a decoder shared by all assets queries it to produce neural Gaussians carrying normals, using the anchor depth to separate the front and back caps.
Both networks are trained jointly, and the decoder produces Gaussian parameters for each camera.
Each asset stores only its anchors and feature planes, whose 2D planar structure makes the representation compatible with standard image and video codecs for efficient compression and deployment. Materials and lighting can then be independently assigned or modified at the receiving end, enabling flexible appearance control.

We construct a paired dataset of 27,221 assets spanning 256 font families and SVG icon collections.
Selected letters, digits, and punctuation, together with icon outlines, are paired with normalized extrusions, anchors, and calibrated normal and silhouette images from 40 views.
Family-disjoint splits support learning and evaluating conversion to unseen typefaces, while the paired geometry and views support matched per-asset fitting comparisons.

On the full test split, \method{} achieves $1.62^\circ$ normal error and $0.985$ silhouette IoU, the best values among the compared feed-forward converters, with a prepared-anchor conversion time of 9.1\,ms per asset.
It also has lower forward computation and peak memory than the compared feed-forward methods.
On a shared 64-asset subset, its one-pass normal error is lower than 3DGS and a single-plane variant of TV-GS fitted per asset even after 32k optimization steps, while Scaffold-GS surpasses it only after 16k steps at roughly triple the uncompressed per-asset storage.
On the same subset, HEVC coding at QP~22 stores an asset in 23.9\,kB on average, raising normal error by $0.05^\circ$.
Our main contributions are:
\begin{itemize}
  \item A feed-forward converter from glyph and SVG geometry to Gaussian assets that preserve the source silhouette and surface normals, enabling physically based shading under user-assigned homogeneous materials and lighting.
  \item A paired dataset spanning font families and SVG outlines, with anchors and calibrated multi-view supervision for conversion and family-level generalization studies.
  \item Experiments demonstrating conversion fidelity and low forward cost against feed-forward and per-asset fitting baselines, with controlled studies of plane prediction, feature layout, refinement, and coded delivery.
\end{itemize}

\section{Related work}
\label{sec:related}

\begin{figure*}[t]
  \centering
  \includegraphics[width=\linewidth]{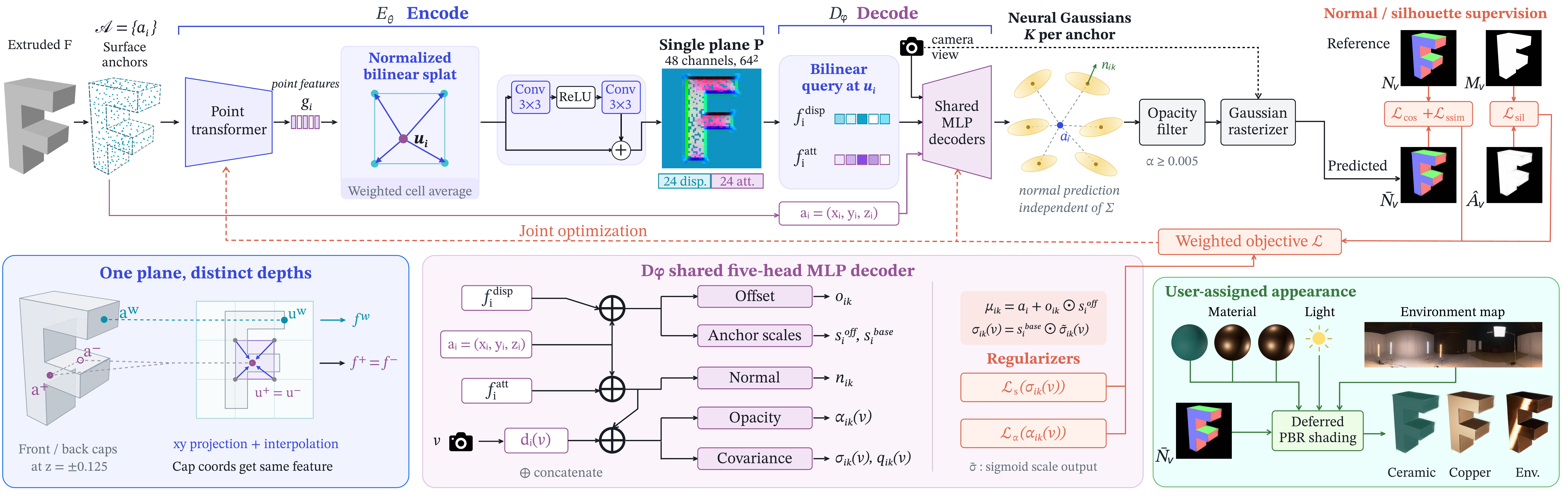}\vspace{0.2em}
  \caption{\textbf{Geometry-to-Gaussian conversion.} Top: the encoder $\enc$ maps surface anchors from the extruded outline to a single $64^2$ feature plane through a point transformer, normalized bilinear splatting and two residual convolutions, and the decoder $\dec$ combines bilinearly queried displacement and attribute features with the anchor coordinate $\mathbf{a}_i$ to predict $K$ neural Gaussians per anchor, which pass an opacity filter and are rasterized for normal and silhouette supervision of both networks. Bottom left: opposite caps project to the same plane coordinate and feature, and the raw depth in $\mathbf{a}_i$ separates them. Bottom center: the shared five-head decoder predicts offsets, anchor scales, normals, opacity, and covariance, where only opacity and covariance depend on the camera direction and normals are independent of covariance. Bottom right: rendered normals feed deferred PBR shading under user-assigned materials and lighting.} %
  \label{fig:pipeline}
\end{figure*}

\noindent\textbf{Gaussian reconstruction and conversion.}
Gaussian representations can be obtained by optimizing an individual asset or by predicting it with a network shared across assets.
3DGS~\cite{kerbl20233dgs} fits free Gaussians to posed images with pruning and adaptive densification, while Scaffold-GS~\cite{lu2024scaffoldgs} decodes Gaussian attributes from spatial anchors and their features.
For known geometry, fitting becomes conversion from a supplied surface.
PBR-TV-GS~\cite{lee2026pbrtvgs} optimizes tri-planes and decoder networks using normals and material values obtained directly from mesh intersections and UV textures.
Mesh2Splat~\cite{scolari2025mesh2splat} and GaMeS~\cite{waczynska2024games} attach Gaussians to mesh triangles and 2DGS~\cite{huang20242dgs} fits surface-aligned disks, each yielding per-asset primitives rather than a fixed-size, codec-ready feature plane.
Feed-forward methods instead predict Gaussians from one or a few views. Splatter Image~\cite{szymanowicz2024splatterimage}, GS-LRM~\cite{zhang2024gslrm} and LGM~\cite{tang2024lgm} use per-pixel outputs, whereas TriplaneGaussian~\cite{zou2024triplanegaussian} predicts through a point cloud and tri-plane.
These methods infer the surface from its images.
Geometry-conditioned approaches receive the surface directly. TexGaussian~\cite{xiong2025texgaussian} predicts material Gaussians on a mesh-derived octree in one pass, and GaussianGrow~\cite{zhang2026gaussiangrow} grows Gaussians from a point cloud through per-asset optimization under a text-guided diffusion prior.
Like these approaches, we condition on supplied geometry, but perform the conversion in a single feed-forward pass, with the rendered normal field and silhouette as the targets to preserve.
Whereas TexGaussian predicts appearance on a surface that remains explicit at render time, our Gaussians must themselves reproduce that surface.

\noindent\textbf{Surface appearance and relighting.}
Normals and reflectance allow Gaussian geometry to support relighting instead of storing only rendered color.
GaussianShader~\cite{jiang2024gaussianshader}, R3DGS~\cite{gao2024relightable3dg} and GS-IR~\cite{liang2024gsir} estimate surface appearance and lighting from multi-view images using Gaussian representations augmented with normals and reflectance.
RelitLRM~\cite{zhang2024relitlrm} combines feed-forward geometry prediction from sparse views with diffusion-based generation of relit Gaussian appearance.
In our setting, the known geometry supplies normal supervision, while the material and lights are assigned by the user.

\noindent\textbf{Structured and compact representations.}
A regular image layout of features makes Gaussian data accessible to image networks and standard codecs.
Splatter Image~\cite{szymanowicz2024splatterimage} stores a Gaussian per pixel, UVGS~\cite{rai2025uvgs} maps a fitted object to a spherical UV image for generative modeling, and SIGMAN~\cite{yang2025sigman} learns a UV-structured latent for human avatars.
Feature planes also build on factorizations used in neural rendering~\cite{chan2022eg3d,chen2022tensorf,fridovichkeil2023kplanes}.
CodecGS~\cite{lee2025codecgs} replaces per-Gaussian attributes with tri-plane features for video coding, and TV-GS~\cite{kathariya2026tv} codes anchor features as video frames alongside an octree of anchor positions.
Other approaches address different parts of delivery. FCGS~\cite{chen2025fcgs} learns to compress fitted Gaussian sets, while AtlasLC~\cite{kim2026atlaslc} compresses object-centric Gaussian assets into codec-ready atlases, treating the reusable asset as the delivery unit.
Our representation adopts TV-GS's anchor and feature-plane layout and learns the mapping from an asset's sampled geometry to its feature plane, queried by a decoder shared by all assets.

\noindent\textbf{Vector graphics, 3D text and training data.}
Generative work on typography and vector graphics addresses the creation and stylization of glyphs and vector art.
Word-As-Image~\cite{iluz2023wordasimage} deforms outlines to convey a word's meaning, VecGlypher~\cite{huang2026vecglypher} generates glyph outlines with a language model, and DeepSVG~\cite{carlier2020deepsvg}, StarVector~\cite{rodriguez2023starvector} and OmniSVG~\cite{yang2025omnisvg} generate SVG graphics.
DreamFont3D~\cite{li2024dreamfont3d} and SplatFont3D~\cite{gan2025splatfont3d} optimize stylized 3D fonts per asset from text prompts.
Our method instead starts from a chosen outline and preserves its extruded geometry.
Existing datasets provide several of the ingredients needed to supervise this task.
DeepCAD~\cite{wu2021deepcad} pairs vector profiles with extrusions, and MV2Cyl~\cite{hong2024mv2cyl} renders them from multiple views.
ABC~\cite{koch2019abc} provides exact normals of CAD models, OmniObject3D~\cite{wu2023omniobject3d} provides multi-view normals of scanned objects, and Syn3DTxt~\cite{hsiung2025syn3dtxt} supplies rendered 3D text for scene-text generation and editing.
However, to our knowledge, no existing dataset jointly provides glyph and SVG outlines, sampled anchors, and calibrated normal/silhouette renderings of their extrusions.
We curate this paired dataset for our conversion task, with a font-family split to evaluate conversion to unseen typefaces.

\section{Method}
\label{sec:method}

\subsection{Problem and representation}
\label{sec:overview}

We convert the extruded \emph{outline} of a glyph or SVG graphic, \ie, its closed 2D contours, into a Gaussian asset whose materials and lighting are assigned afterwards (\cref{fig:pipeline}).
The input is a set of normalized anchors $\anchors=\{\mathbf{a}_i\}_{i=1}^{M}$, with $\mathbf{a}_i=(x_i,y_i,z_i)$, sampled from the known surface during preprocessing.
In \method{}, an asset is represented by its anchors and a feature plane from which a decoder predicts Gaussians.
An encoder $\enc$ maps the anchor positions to a single 2D feature plane $\plane\in\mathbb{R}^{C\times G\times G}$ with $C=48$ channels. $G=64$ is the side dimension of the feature plane. 
A decoder $\dec$ reads $\plane$ at each anchor and predicts $K=6$ candidate Gaussians carrying surface normals.
On our dataset, the encoder and decoder are trained end-to-end from normals and silhouette of the source geometry.

As shown in \cref{fig:pipeline}, a new asset needs one feed-forward encoder pass, $\plane=\enc(\anchors)$, \ie, no per-asset optimization. The decoder is then queried for each camera.
Each asset stores only its anchors and plane, while the decoder is stored once and shared by all assets.
A single plane, rather than the tri-plane of TV-GS~\cite{kathariya2026tv}, suffices because an extrusion is determined by its outline and depth. The outline defines the plane's domain and the anchor depth given to the decoder separates the front and back caps, which share plane coordinates.

\subsection{Geometry preprocessing}
\label{sec:anchors}

Normal error concentrates at the rim, where the caps meet the side walls and the surface normal changes abruptly, yet this narrow region makes up little of the surface area.
Preprocessing therefore concentrates anchors there.
We first normalize and rasterize the outline within $[-1,1]^2$, extract simplified polygonal contours, and extrude the enclosed region to $z=\pm0.125$.
We then place 35\% of anchors along rim edges, favoring high-curvature contour regions, and distribute the rest by farthest-point sampling over the surface.
The anchor count $M$ grows with contour length at fixed extrusion depth, bounded by $1000\le M\le4096$, with sampling constants fixed across the dataset.
Only normalized anchor coordinates enter the network as input.

\subsection{Encoding geometry in a feature plane}
\label{sec:plane}

Anchor counts range from 1000 to 4096 across assets, whereas compact delivery with standard codecs needs a latent of fixed size and layout.
We therefore predict a single image-shaped feature plane from the anchors, replacing the per-asset plane fitting of TV-GS~\cite{kathariya2026tv} with one encoder pass.

A point transformer~\cite{zhao2021pointtransformer} extracts local context in six vector-attention layers over the 16 nearest neighbors of each anchor $\mathbf{a}_i$.
Two linear layers map its features $\mathbf{h}_i\in\mathbb{R}^{96}$ to $\mathbf{g}_i\in\mathbb{R}^{C}$.
These per-anchor features are then written into the $G\!\times\! G$ plane.
Anchor $i$ has plane coordinate
\begin{equation}
  \mathbf{u}_i=\rho\,\big((x_i,y_i)-\mathbf{c}_{\anchors}\big)\in[-1,1]^2 ,
  \label{eq:planecoord}
\end{equation}
where $\mathbf{c}_{\anchors}$ is the midpoint of the anchors' bounding box in the outline plane.
The scale $\rho$ makes the longer side of this bounding box span the plane, minus a one-cell border, and is capped at 3 so that small shapes such as punctuation marks are not overly magnified.

Each feature $\mathbf{g}_i$ is spread over the $2\times2$ cells whose centers surround $\mathbf{u}_i$, with bilinear weights $w_{ic}$ that grow as $\mathbf{u}_i$ approaches the center of cell $c$.
Each cell averages what it receives, $\plane^{0}_{c}=\sum_i w_{ic}\mathbf{g}_i/\sum_i w_{ic}$, and cells that receive no weight are set to $\mathbf{0}$.
Weight normalization prevents densely sampled regions, such as rims, from producing larger feature magnitudes simply through more contributions.
Additionally, two $3\times3$ convolutions with a residual connection exchange information between neighboring cells to produce $\plane$.
We divide its channels equally into \emph{displacement} features $\plane^{\mathrm{disp}}$ and \emph{attribute} features $\plane^{\mathrm{att}}$.
For decoding, both halves are read back at $\mathbf{u}_i$ by bilinear interpolation, giving $\mathbf{f}_i^{\mathrm{disp}}$ and $\mathbf{f}_i^{\mathrm{att}}$.

\subsection{Decoding neural Gaussians}
\label{sec:decoder}

The decoder must recover a closed 3D surface from features stored on a 2D plane.
Points on the front and back caps with the same $(x,y)$ receive the same plane feature, although their normals point in opposite directions.
Each decoder head therefore also receives the original anchor coordinate, including raw depth $z$, to separate them.
We adapt TV-GS's five-head, two-layer MLP decoder~\cite{kathariya2026tv} to predict the $K$ Gaussians per anchor, with normals replacing color channels (Supp.~Tab.~\ref{tab:supp-heads}).

Two placement heads use displacement features to predict offsets $\mathbf{o}_{ik}\in\mathbb{R}^3$ and per-anchor scales $\mathbf{s}_i^{\mathrm{off}},\mathbf{s}_i^{\mathrm{base}}\in\mathbb{R}^3$.
Gaussian means remain in the original anchor frame:
\begin{equation}
  \boldsymbol{\mu}_{ik}=\mathbf{a}_i+\mathbf{o}_{ik}\odot\mathbf{s}_i^{\mathrm{off}},\qquad k=1,\dots,K,
  \label{eq:means}
\end{equation}
where $\odot$ denotes the elementwise product.
The remaining three heads use attribute features to predict normals, opacity, and covariance.
The normal head predicts unit-length $\mathbf{n}_{ik}$ independently of covariance, separating surface orientation from Gaussian extent near sharp rims.

Only opacity and covariance additionally depend on the unit direction $\mathbf{d}_i(v)$ from camera $v$ to the anchor.
Opacity $\alpha_{ik}(v)$ is a sigmoid output and covariance uses scales $\boldsymbol{\sigma}_{ik}(v)$ formed by multiplying $\mathbf{s}_i^{\mathrm{base}}$ elementwise with a sigmoid output, and rotations $\mathbf{q}_{ik}(v)$ represented by normalized quaternions.
Opacity filtering discards candidates with $\alpha_{ik}(v)<0.005$ from the decoded Gaussians. During training, anchors stay fixed with no pruning or densification.

\begin{table*}[t]
\centering\footnotesize
\setlength{\tabcolsep}{6pt}
\caption{\textbf{Conversion fidelity on held-out views.} Asset-averaged normal error over foreground pixels and their edge band, $\Delta$ normal relative to \method{} on the same population, and silhouette IoU. Retrained feed-forward methods receive the listed inputs, while fitting methods optimize against 32 available views per asset for the stated step budgets. \bestkey{Red} and \secondkey{orange} cells mark the best and second-best values per column within each block. The $\Delta$ column is relative and left unmarked.}
\label{tab:comparison}
\begin{tabular}{@{}llp{6em}p{8em}p{6em}p{6em}}
\toprule
Method & Input & Normal ($^\circ$) $\downarrow$ & Edge-band ($^\circ$) $\downarrow$ & $\Delta$ normal & IoU $\uparrow$ \\
\midrule
\multicolumn{6}{@{}l}{\textit{Full test split: 2,657 assets}} \\
TexGaussian~\cite{xiong2025texgaussian} & Mesh $\to$ points/normals & \second 1.957 & \second 13.86 & +0.339 & 0.9595 \\
TriplaneGaussian~\cite{zou2024triplanegaussian} & Anchors + 1 view & 5.926 & 18.84 & +4.308 & \second 0.9810 \\
LGM~\cite{tang2024lgm} & 4 views & 12.873 & 32.51 & +11.255 & 0.9576 \\
\textbf{\method{}} & Anchors & \best 1.618 & \best 8.98 & -- & \best 0.9851 \\
\midrule
\multicolumn{6}{@{}l}{\textit{64-asset fitting subset}} \\
3DGS~\cite{kerbl20233dgs}, 500 & Anchors + available views & 5.185 & 21.95 & +3.484 & 0.8357 \\
3DGS, 2k & Anchors + available views & 2.123 & 14.74 & +0.422 & 0.9405 \\
3DGS, 32k & Anchors + available views & 2.099 & 13.97 & +0.398 & 0.9811 \\
Scaffold-GS~\cite{lu2024scaffoldgs}, 2k & Anchors + available views & 2.733 & 16.55 & +1.032 & 0.9400 \\
Scaffold-GS, 32k & Anchors + available views & \second 1.514 & 10.95 & $-$0.187 & \best 0.9863 \\
Plane-and-decoder fitting, 32k & Anchors + available views & 1.897 & 13.35 & +0.196 & 0.9828 \\
Mesh2Splat~\cite{scolari2025mesh2splat}, default & Mesh & 2.371 & 20.89 & +0.670 & 0.9535 \\
GaMeS~\cite{waczynska2024games}, 32k & Mesh + available views & 2.764 & 15.71 & +1.063 & 0.9700 \\
\textbf{\method{}} & Anchors & 1.701 & \best 8.90 & -- & \second 0.9840 \\
\textbf{\method{} + 10\,s refinement} & Anchors + available views & \best 1.313 & \second 8.98 & $-$0.388 & 0.9825 \\
\bottomrule
\end{tabular}
\end{table*}

\subsection{Rendering and supervision}
\label{sec:shading}

A converted asset has no captured photographs to match, and its material and lighting are chosen only at use time.
We therefore supervise geometry alone, comparing normals and silhouettes rendered from the Gaussians with those of the source geometry.

\noindent\textbf{Normal rendering.}
Let $\mathbf{N}_v$ and $\mathbf{M}_v$ be the reference world-space normal image and silhouette at view $v$.
We rasterize Gaussians with normals in place of colors~\cite{kerbl20233dgs,ye2025gsplat}.
At pixel $p$, index the covering Gaussians in depth order by $j$, with projected opacity $\alpha_j$ and transmittance $T_j=\prod_{j'<j}(1-\alpha_{j'})$.
The rendered normal and alpha are $\hat{\mathbf{N}}_v(p)=\sum_j \mathbf{n}_j\alpha_j T_j$ and $\hat{\mathbf{A}}_v(p)=\sum_j \alpha_j T_j$.
We normalize the accumulated normal, $\bar{\mathbf{N}}_v(p)=\mathrm{normalize}(\hat{\mathbf{N}}_v(p))$, before scoring normal error or shading.

\noindent\textbf{Training objective.}
Each step samples 32 assets and 16 views per asset from the training views.
Training minimizes
\begin{equation}
  \mathcal{L}=\lambda_{\cos}\mathcal{L}_{\cos}+\lambda_{\mathrm{ssim}}\mathcal{L}_{\mathrm{ssim}}+\lambda_{\mathrm{sil}}\mathcal{L}_{\mathrm{sil}}
  +\lambda_{\mathrm{s}}\mathcal{L}_{\mathrm{s}}+\lambda_{\alpha}\mathcal{L}_{\alpha}.
  \label{eq:loss}
\end{equation}
The normal terms compare pixel orientations and local image structure.
Let $\Omega_v$ contain foreground pixels with valid reference normals, excluding partially covered boundary pixels, and $\mathbb{1}_{\Omega_v}$ be its indicator image.
Then
\begin{equation}
\begin{aligned}
  \mathcal{L}_{\cos}&=\frac{1}{|\Omega_v|}\sum_{p\in\Omega_v}\Big(1-\big\langle\bar{\mathbf{N}}_v(p),\mathbf{N}_v(p)\big\rangle\Big),\\
  \mathcal{L}_{\mathrm{ssim}}&=1-\mathrm{SSIM}\big(\bar{\mathbf{N}}_v\odot\mathbb{1}_{\Omega_v},\ \mathbf{N}_v\odot\mathbb{1}_{\Omega_v}\big),
\end{aligned}
  \label{eq:normalloss}
\end{equation}
with SSIM~\cite{wang2004ssim} evaluated on masked normal maps.
The silhouette loss,
$\mathcal{L}_{\mathrm{sil}}=\operatorname{mean}_{p}|\hat{\mathbf{A}}_v(p)-\mathbf{M}_v(p)|$,
covers all pixels, penalizing both missing foreground and opacity outside the shape.
The regularizers $\mathcal{L}_{\mathrm{s}}$ and $\mathcal{L}_{\alpha}$ average $\lVert\boldsymbol{\sigma}_{ik}(v)\rVert_2$ and Gaussian opacity over the Gaussians passing the opacity filter, discouraging large extents and opacity.
Per-view image losses and regularizers are averaged over each asset's sampled views, then over assets.
We use $(\lambda_{\cos},\lambda_{\mathrm{ssim}},\lambda_{\mathrm{sil}},\lambda_{\mathrm{s}},\lambda_{\alpha})=(20,1,1,5\times10^{-3},10^{-2})$.

\subsection{Refinement and appearance control}
\label{sec:refinement}

For plane refinement, we initialize $\plane$ from $\enc(\anchors)$, freeze the decoder, and minimize \cref{eq:loss} over the plane alone using Adam~\cite{kingma2015adam} and reference images of the known geometry.
We store the optimized plane with unchanged dimensions and 16-bit payload, using the same quantization and codec as for predictions.
Our main refinement budget is 10\,s. Optimization and timing details are in Supp.~Sec.~\ref{supp:refinement}.

At use time, each asset receives a homogeneous material, including albedo, roughness, and metallicity, and lights or an environment map.
Following PBR-TV-GS~\cite{lee2026pbrtvgs}, we apply deferred metallic-roughness Cook--Torrance shading~\cite{cook1982reflectance} to the rasterized normal and alpha images, so appearance can change without reconverting geometry (Supp.~Sec.~\ref{supp:appearance}).

\section{Experiments}
\label{sec:experiments}

\subsection{Experimental setup}
\label{sec:setup}

\noindent\textbf{Dataset.}
Our dataset contains 27,221 glyph and SVG icon assets, each pairing an outline with its normalized extrusion, sampled anchors, and world-space normal and silhouette images rendered by Blender~\cite{blender} from 40 cameras at $512^2$ resolution.
We split glyphs by font family, with 206 families for training and 25 each for validation and testing. All characters occur in every split.
Icons are split by source collection.
The splits contain 21,904 training, 2,660 validation, and 2,657 test assets. Construction details and statistics are in Supp.~Sec.~\ref{supp:dataset}.

\noindent\textbf{Evaluation.}
Of the 40 cameras, 32 provide views for training, fitting, and refinement. The remaining eight are held out for scoring every method.
We report \emph{normal error}, the mean angular difference between rendered and reference world-space normals over foreground pixels, and \emph{silhouette IoU}, thresholding rendered alpha at $0.5$.
The \emph{edge band} contains foreground pixels within one pixel of a reference-normal crease of at least $30^\circ$.
Its share of foreground pixels over all 40 reference views defines \emph{edge fraction}, a model-free measure of geometric difficulty used to select and stratify assets.
All outputs use the same rasterizer~\cite{ye2025gsplat} at $512^2$. Scores are averaged over the eight views per asset, then over assets.
Full-test results average three training seeds for every converter, and the 64-asset fitting comparison uses one.
Hyperparameters are selected on the validation assets. %
Forward cost covers prepared inputs to Gaussian parameters at batch size 1 on an NVIDIA RTX A6000.

\subsection{Fidelity and cost of one-pass conversion}
\label{sec:main}

\Cref{tab:comparison} compares one-pass conversion on the full test split and per-asset fitting on a shared 64-asset subset.

\noindent\textbf{Feed-forward conversion.}
TexGaussian~\cite{xiong2025texgaussian} receives mesh-sampled points and normals, TriplaneGaussian~\cite{zou2024triplanegaussian} receives anchors and one normal view, and LGM~\cite{tang2024lgm} receives four normal views.
All are retrained on our training split with normal heads replacing color heads and rendered normals as the main supervision signal.
\method{} receives anchor positions alone.

On the 2,657 test assets, \method{} has the lowest normal error ($1.62^\circ$) and highest IoU ($0.985$) among the compared converters, improving on TexGaussian by $0.34^\circ$ and $0.026$, respectively.
The normal-error gap widens at edges, $8.98^\circ$ versus TexGaussian's $13.86^\circ$ and $18.84^\circ$ for TriplaneGaussian, whose closer IoU ($0.981$), despite $5.93^\circ$ normal error, shows that similar silhouettes can conceal errors in surface orientation.
\method{} also has the lowest forward cost, 9.1\,ms and 97\,MiB per asset versus 28.7--157.9\,ms and 1,400--2,843\,MiB for the others (\cref{tab:conversion-resources}).

\begin{table}[t]
\centering\small
\setlength{\tabcolsep}{3pt}
\caption{\textbf{Forward conversion cost.} Time, throughput and peak memory per asset from prepared inputs to Gaussian parameters, batch one, on a single RTX A6000. \bestkey{Red} and \secondkey{orange} cells mark the best and second-best values.}
\label{tab:conversion-resources}
\begin{tabular}{@{}lrrr@{}}
\toprule
Method & Forward (ms)$\downarrow$ & Assets/s$\uparrow$ & Peak GPU (MiB)$\downarrow$ \\
\midrule
TexGaussian & 157.9 & 6.3 & 1,675 \\
TriplaneGaussian & \second 28.7 & \second 34.9 & \second 1,400 \\
LGM & 47.9 & 20.9 & 2,843 \\
\textbf{\method{}} & \best 9.1 & \best 109.9 & \best 97 \\
\bottomrule
\end{tabular}
\end{table}

\noindent\textbf{Per-asset fitting.}
We fit 3DGS~\cite{kerbl20233dgs}, Scaffold-GS~\cite{lu2024scaffoldgs}, and our representation from the same anchors and 32 available normal and silhouette views.
The last baseline, \emph{plane-and-decoder fitting}, adapts TV-GS~\cite{kathariya2026tv} to a single plane, optimizing the plane and decoder from scratch per asset.
It tests whether fitting the same representation can match amortized prediction.
All methods use the same test assets, selected at evenly spaced ranks of edge fraction. Adaptations and the complete fitting trajectories are in Supp.~Sec.~\ref{supp:baselines}.

In one forward pass, \method{} reaches $1.70^\circ$ normal error, below 3DGS and plane-and-decoder fitting at every budget up to 32k steps, and it has the lowest edge-band error of all methods at every budget: $8.90^\circ$ versus $10.95^\circ$ for Scaffold-GS and $13.35^\circ$--$15.71^\circ$ for the other fits at 32k.
Only Scaffold-GS reaches a lower overall error, $1.46^\circ$ after 16k steps, with slightly higher IoU at 16k and 32k.
It relies on a denser representation: at 32k, the median Scaffold-GS asset has 8,564 anchors and 1.23\,MB of half-precision parameters, versus 1,337 anchors and 401\,kB for \method{}, plus a 45.5\,kB decoder shared by all assets.
Mesh-attached Gaussians are less accurate despite using 22.5\,MB (Mesh2Splat~\cite{scolari2025mesh2splat}) and 4.7\,MB (GaMeS~\cite{waczynska2024games}) per asset (see Supp.~Sec.~\ref{supp:mesh-attached}).
Refining our predicted plane for 10\,s lowers normal error to $1.31^\circ$, below every fitting result, and \cref{sec:refine-exp} examines the regional gains and silhouette trade-off.

\begin{figure*}[!htbp]
  \centering
  \includegraphics[width=\linewidth]{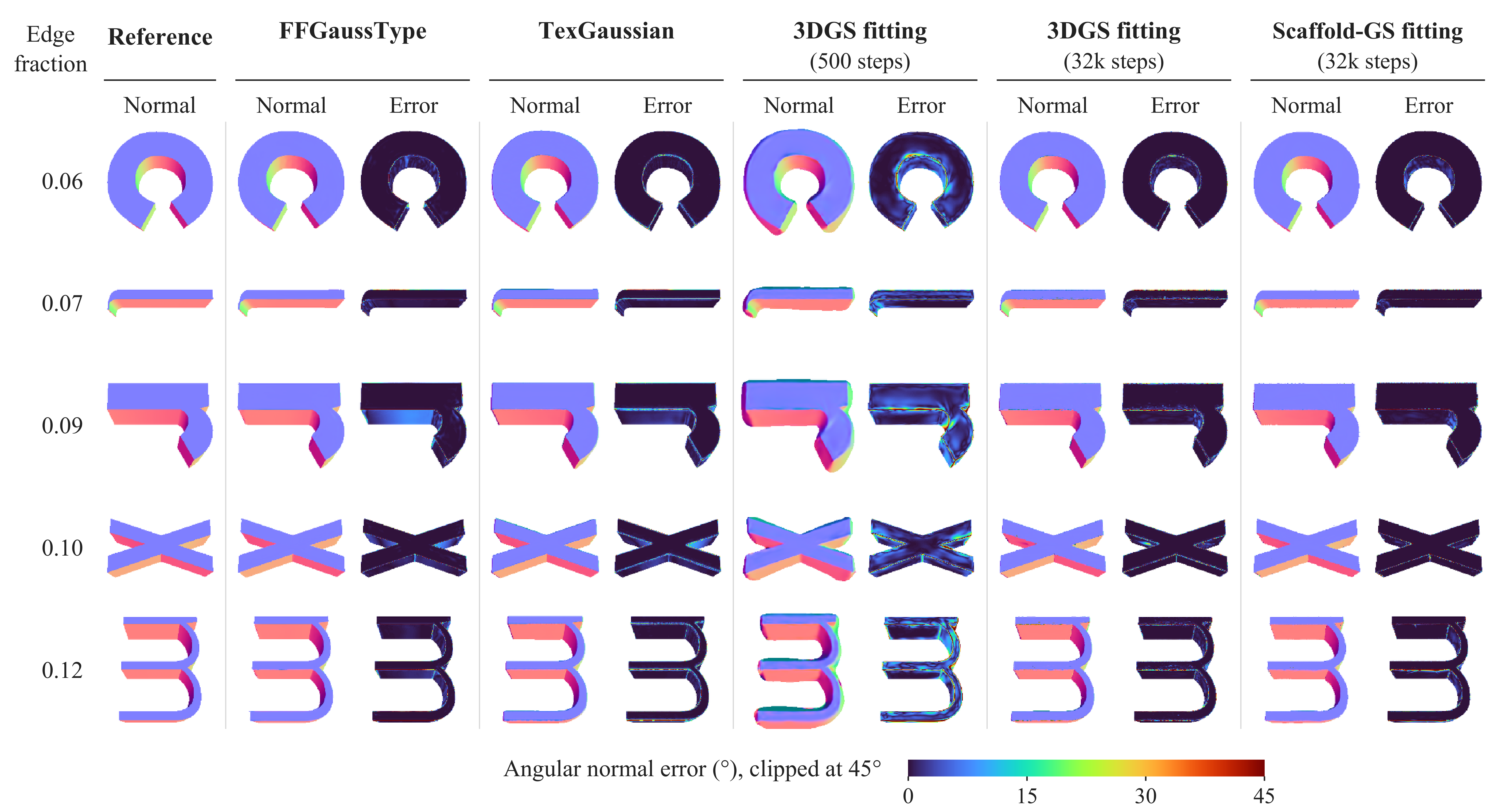}
  \caption{\textbf{Qualitative comparison on five test assets.} Rows follow increasing edge fraction (left labels), using the fitting subset's selection rule and one held-out view per asset. The first column shows reference normals, and subsequent columns pair method normals with angular error clipped at $45^\circ$. RGB encodes world-space normal components $(x,y,z)$, while the color bar applies only to the error maps.}
  \label{fig:qualitative}
\end{figure*}

\Cref{fig:qualitative} shows five test assets spanning the edge-fraction range.
For \method{} and the 32k-step fits, front-face normals are nearly error-free, with residual errors concentrated on narrow side walls and creases.
TexGaussian and short-budget fitting leave errors across whole walls and round corners.
One-pass conversion thus recovers the faces and wall structure of long fits, while finer features remain challenging.

\subsection{Generalization to unseen font families}
\label{sec:family}

\begin{figure}[!t]
  \centering
  \setlength{\unitlength}{\columnwidth}%
  \includegraphics[width=\columnwidth]{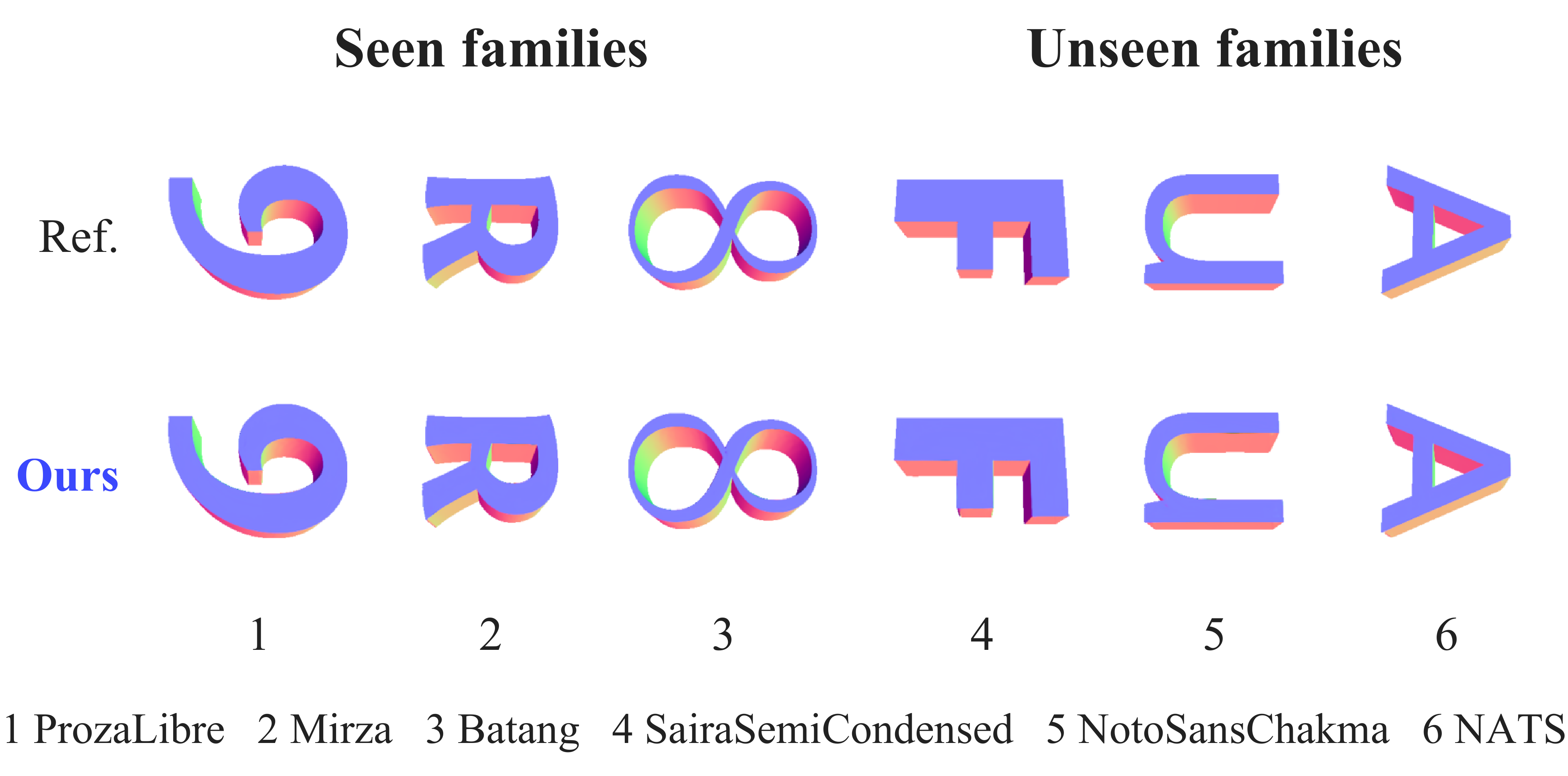}%
  \begin{picture}(0,0)
    \put(-0.4548,0.0743){{\color{gray}\linethickness{0.5pt}\line(0,1){0.4011}}}
  \end{picture}
  \caption{\textbf{Seen and unseen font families.} Reference and predicted world-space normals for three training (seen) and three test (unseen) font families at held-out views.}
  \label{fig:family}
\end{figure}

The family-disjoint split tests familiar characters in unseen typefaces.
Unseen families have $1.63^\circ$ normal error versus $1.65^\circ$ for seen families, with IoU of $0.985$ for both.
Unseen families score marginally better, and the $0.02^\circ$ gap is small relative to the $0.37^\circ$ interquartile range across families, supporting similar average fidelity on seen and unseen typefaces.
\Cref{fig:family} shows paired reference and predicted normals, and Supp.~Sec.~\ref{supp:family} provides sensitivity analyses.

\subsection{Ablation studies}
\label{sec:ablation}

We study the prediction and representation choices that let a single feature plane describe the extruded geometry.

\begin{figure*}[tb]
  \centering
  \includegraphics[width=\linewidth]{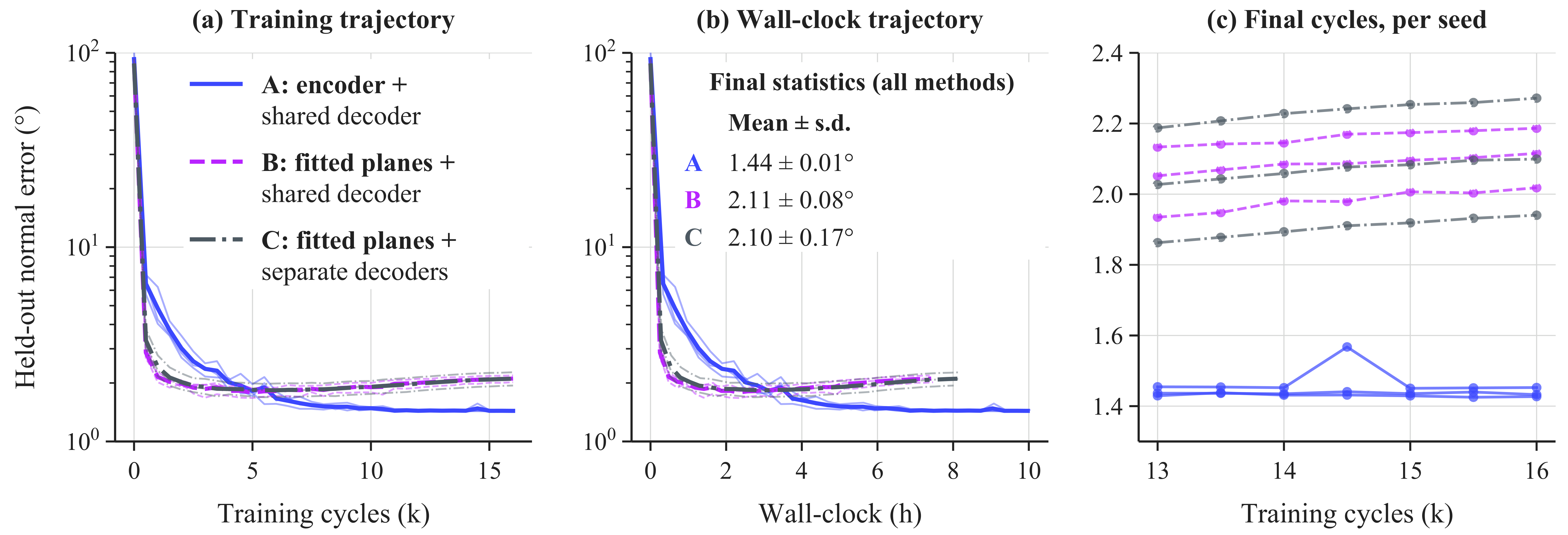}
  \caption{\textbf{Plane prediction and decoder sharing.} Held-out normal error on 32 training assets over three seeds, against training cycles and wall-clock time, with per-seed errors over the final cycles. Statistics in (b) give the mean $\pm$ s.d. over three seeds at the last checkpoint. A predicts planes with a shared decoder, while B and C fit planes with shared and separate decoders, respectively.}
  \label{fig:study}
\end{figure*}

\noindent\textbf{Plane prediction and decoder sharing.}
On 32 training assets, we compare encoder-predicted planes with a shared decoder (A), directly fitted planes with a shared decoder (B), and directly fitted planes with separate decoders (C).
Each setting uses three seeds and 16k training cycles with matched view exposure (Supp.~Sec.~\ref{supp:study}), scored on the same eight held-out views per asset.
\Cref{fig:study} shows that A reaches $1.44^\circ$ normal error, compared with $2.11^\circ$ for B and $2.10^\circ$ for C, and improves on B for every asset.
Fitted planes reach their lowest held-out error after 5k--6k cycles and then degrade, whereas the encoder continues to improve.
Prediction ties all planes to a common mapping from anchor geometry instead of letting each specialize to its fitting views, consistent with a regularizing effect, and the nearly identical errors of B and C show that decoder sharing alone does not explain the gain.

\noindent\textbf{Latent layout.}
\Cref{tab:architecture} compares latent layouts on the validation split under the same training recipe.
Per-anchor features derived from each $\mathbf{h}_i$ without spatial aggregation give the lowest mean normal error, $1.58^\circ$ versus $1.67^\circ$ for the single plane. The image-shaped latent trades some normal fidelity for a fixed spatial layout, the property TV-GS exploits for delivery with standard video codecs.
The tri-plane controls concatenate features from the $xy$, $xz$ and $yz$ planes and approximately match the single plane's latent budget (Supp.~Sec.~\ref{supp:encoder}).
Neither allocation improves on the single plane. The $64\times64$ tri-plane with fewer channels per plane matches it within the seed spread, and the full-channel $37\times37$ control degrades both metrics.
Further controls on plane resolution, the Octree-CNN encoder, and training length are reported in Supp.~Sec.~\ref{supp:ablations}.

\begin{table}[t]
\centering\footnotesize
\setlength{\tabcolsep}{4pt}
\caption{\textbf{Latent layout and depth feature.} Validation normal error and silhouette IoU, reported as means $\!\pm\!$ standard deviations over 3 training seeds. Latent size is the raw 16-bit plane payload.}
\label{tab:architecture}
\begin{tabular}{@{}lrrr@{}}
\toprule
Variant & Latent (kB) & Normal ($^\circ$) & IoU \\
\midrule
Per-anchor features & -- & 1.581\,{\scriptsize$\pm$0.029} & 0.983\,{\scriptsize$\pm$0.0016} \\
Tri-plane, fewer channels & 393.2 & 1.683\,{\scriptsize$\pm$0.045} & 0.982\,{\scriptsize$\pm$0.0018} \\
Tri-plane, lower resolution & 394.3 & 1.876\,{\scriptsize$\pm$0.162} & 0.978\,{\scriptsize$\pm$0.0031} \\
Plane, no depth & 393.2 & 1.757\,{\scriptsize$\pm$0.070} & 0.975\,{\scriptsize$\pm$0.0024} \\
Plane + $\gamma(z)$ & 393.2 & 1.675\,{\scriptsize$\pm$0.132} & 0.984\,{\scriptsize$\pm$0.0010} \\
\textbf{Plane + raw $z$ (ours)} & 393.2 & 1.668\,{\scriptsize$\pm$0.057} & 0.984\,{\scriptsize$\pm$0.0020} \\
\bottomrule
\end{tabular}
\end{table}

\noindent\textbf{Depth feature.}
\method{} feeds the raw anchor depth to the input of every head as part of the full anchor coordinate.
Removing this channel lowers IoU from $0.984$ to $0.975$ and raises normal error from $1.67^\circ$ to $1.76^\circ$, so the interpolated plane feature alone does not separate the two caps.
Replacing the scalar with a four-channel sinusoidal encoding $\gamma(z)$ (Supp.~Sec.~\ref{supp:ablations}) gives $1.68^\circ$ and the same IoU. Therefore, we keep the raw depth input for simplicity.

\subsection{Plane refinement}
\label{sec:refine-exp}

\Cref{tab:refinement} evaluates 1, 10 and 60\,s of plane refinement on the 64-asset subset with the shared networks fixed.
At 10\,s, normal error falls from $1.70^\circ$ to $1.31^\circ$, improving on 62 of the 64 assets, while IoU drops from $0.984$ to $0.982$.
The gain is concentrated in flat regions ($1.00^\circ$ to $0.57^\circ$), with much of it already present at 1\,s. Edge error stays near $8.7$--$9.0^\circ$.

\begin{table}[t]
\centering\footnotesize
\setlength{\tabcolsep}{3.7pt}
\caption{\textbf{Refinement on the 64-asset subset.} Held-out normal error over all foreground pixels, flat regions and the edge band, and silhouette IoU, before and after 1, 10 and 60\,s. Plane and per-anchor features use separately selected step sizes. \bestkey{Red} and \secondkey{orange} cells mark the best and second-best distinct values.}
\label{tab:refinement}
\begin{tabular}{@{}llrrrr}
\toprule
Latent & Budget & Normal ($^\circ$)$\downarrow$ & Flat ($^\circ$)$\downarrow$ & Edge-band ($^\circ$)$\downarrow$ & IoU$\uparrow$ \\
\midrule
\multirow{4}{*}{\shortstack{Plane\\(ours)}} & none & 1.70 & 1.00 & \second 8.90 & \best 0.984 \\
 & 1\,s & 1.36 & 0.65 & \best 8.69 & \best 0.984 \\
 & 10\,s & \best 1.31 & \second 0.57 & 8.98 & \second 0.982 \\
 & 60\,s & \second 1.35 & \best 0.56 & 9.37 & 0.981 \\
\midrule
\multirow{4}{*}{\shortstack{Per-anchor\\features}} & none & 1.61 & 0.98 & \second 8.12 & \best 0.984 \\
 & 1\,s & 1.21 & \second 0.55 & \best 8.00 & \second 0.983 \\
 & 10\,s & \best 1.16 & \best 0.48 & 8.14 & \second 0.983 \\
 & 60\,s & \second 1.19 & \best 0.48 & 8.52 & 0.982 \\
\bottomrule
\end{tabular}
\end{table}

Extending refinement to 60\,s leaves flat regions nearly unchanged while edges, overall error, and IoU all worsen.
Similarly, per-anchor features, refined with their own validation-selected step size, improve and then deteriorate.
Supp.~Sec.~\ref{supp:refinement} gives the optimization and timing protocol.

\begin{figure}[!t]
  \centering
  \includegraphics[width=\columnwidth]{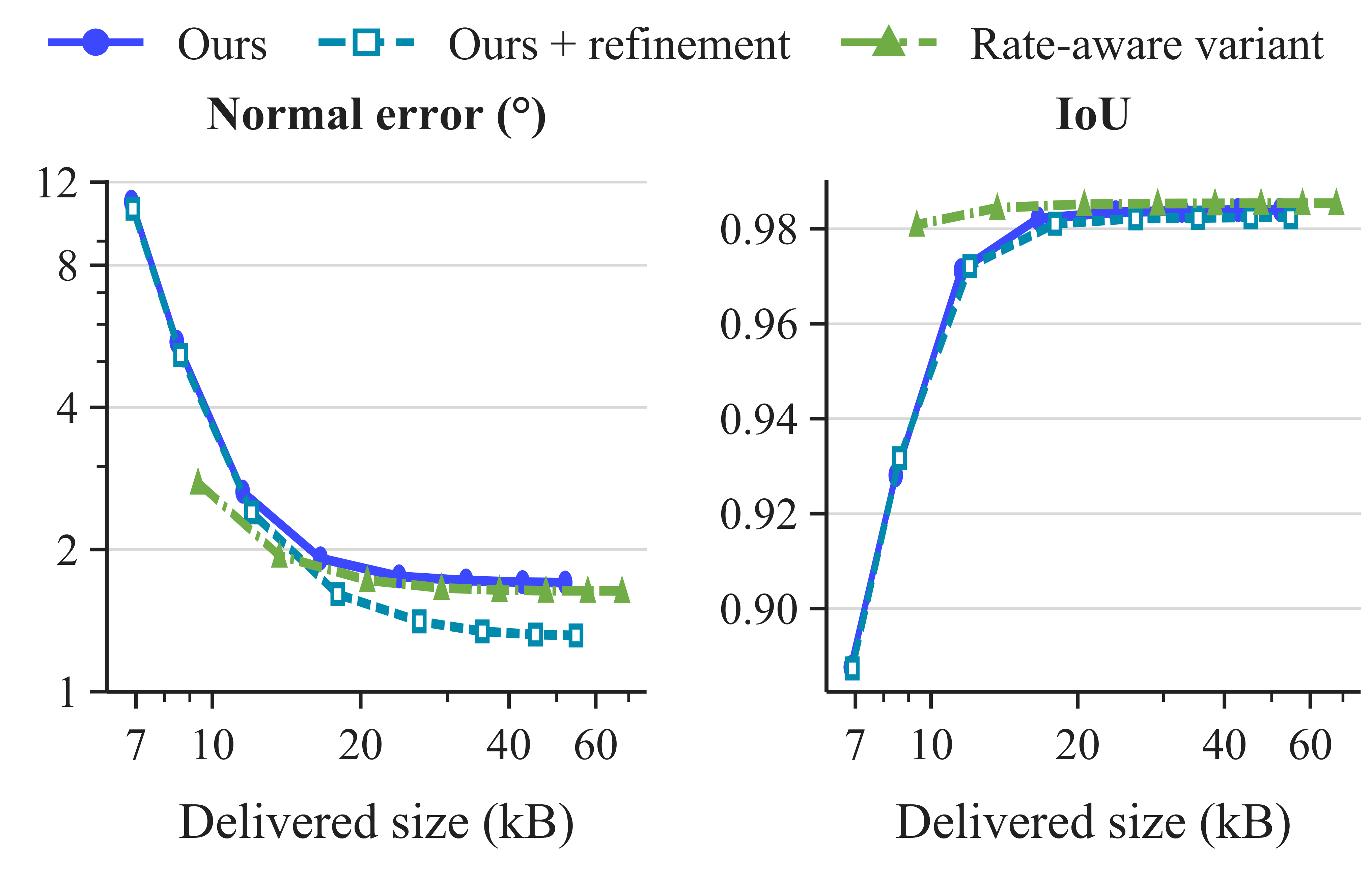}
  \caption{\textbf{Coded delivery on 64 test assets.} Mean held-out normal error and silhouette IoU versus combined anchor-and-plane size over eight HEVC QPs, for default predictions, 10\,s refined planes and the rate-aware variant.}
  \label{fig:delivery}
\end{figure}

\subsection{Asset delivery and reuse}
\label{sec:rd}

We measure how coding affects predicted planes on the 64-asset fitting subset.
The default model uses no compression objective. Its planes are quantized and coded with HEVC~\cite{sullivan2012hevc} after training.
\Cref{fig:delivery} compares normal error and silhouette IoU against the combined size of coded anchors and planes.
Lossless still-image codecs preserve the ten-bit plane codes exactly, and JPEG~XL~\cite{alakuijala2019jpegxl} stores an asset in 72.4\,kB.
With HEVC, fidelity changes little down to about 24\,kB per asset. At QP~22, an asset occupies 23.9\,kB on average, with a $0.05^\circ$ increase in normal error over the uncoded result.
Stronger compression increasingly reduces fidelity.
A further 10\,s of plane refinement improves normal fidelity, storing the optimized plane while keeping the shared networks fixed.
Refined planes cost more plane bytes at equal QP, but retain lower normal error at equal total asset size down to 16.6\,kB.
An optional \emph{rate-aware} variant, trained with the compression objective of TV-GS under its own validation-selected quantizer, is shown for reference.
Supp.~Sec.~\ref{supp:delivery} gives the coding settings and corresponding results (Tabs.~\ref{tab:supp-hevc}, ~\ref{tab:supp-stillimage}), in which anchor quantization changes normal error by at most $0.004^\circ$ and losses come mainly from the plane coding.

Delivered instances can share stored anchors and a feature plane while varying placement, orientation, and appearance independently.
Camera-conditioned decoding and depth ordering provide a basis for composition with Gaussian scenes. In our insertions into volumetric video (\cref{fig:teaser}), the view-conditioned decoding remains visually stable under camera motion and asset animation (see the supplementary videos on the project page).

\section{Conclusion}
\label{sec:conclusion}

We presented \method{}, which converts known glyph and SVG geometry into reusable Gaussian assets instead of fitting each new asset separately.
Trained on paired outlines, sampled anchors, and multi-view renderings, the converter maps anchor positions to a feature plane that a decoder shared by all assets turns into neural Gaussians with surface normals.
It reaches $1.62^\circ$ normal error on the full test split in 9.1\,ms from prepared anchors, the best fidelity among the compared feed-forward converters, and 10\,s of plane refinement surpasses the best per-asset fitting result in normal accuracy.
The plane remains an image-shaped latent that can be refined or coded after prediction, and normal-based rendering leaves materials and lighting configurations to the user.
Our study focuses on extruded glyph and SVG assets with a homogeneous material per asset, and anchor sampling remains a separate preprocessing step performed once per outline.
The adopted shading does not model self-shadowing or light transport between an asset and its surrounding scene.
Supporting spatially varying materials, such as texture mapping, is left to future work.

{\small\bibliographystyle{ieeenat_fullname}\bibliography{refs}}

\clearpage

\appendix
\setcounter{table}{0}\setcounter{figure}{0}
\renewcommand{\thetable}{\thesection\arabic{table}}
\renewcommand{\thefigure}{\thesection\arabic{figure}}
\makeatletter\@addtoreset{table}{section}\@addtoreset{figure}{section}\makeatother
\section{Dataset construction and camera protocol}
\label{supp:dataset}

\noindent\textbf{Geometry and anchors.}
Each asset pairs a source outline with its extruded geometry, surface anchors,
and calibrated normal and silhouette images. Glyphs cover 77 printable ASCII
characters: uppercase and lowercase letters, digits, and 15 punctuation marks.
We rasterize each font or SVG outline at $2048^2$ in a canonical frame,
normalized to about unit extent by the font's em square or the SVG view box,
extract simplified polygonal contours, and extrude the
enclosed region to $z=\pm0.125$ without beveling. These normalized coordinates
are what the network receives.

The anchor budget follows the side-wall area $A_{\mathrm{wall}}$ of the
normalized extrusion. On a sample of this corpus, we searched for suitable
anchor counts and fitted a linear relationship to this area, giving the target
\begin{equation}
 M=\min\!\left(4096,\max\!\left(1000,
       \operatorname{round}(1824.766 A_{\mathrm{wall}})\right)\right).
\end{equation}
This rule is calibrated for this dataset and extrusion depth. Rim-biased sampling and
farthest-point sampling then distribute the anchors as described in the main
paper.
The prepared anchors are shared by all anchor-based methods. The extrusion mesh is
needed only to render supervision, so anchors for a new outline follow from its
2D contours and the fixed extrusion depth.

\noindent\textbf{Selection and splits.}
We retain 27,221 of the 27,742 source assets after quality checks and duplicate
exclusion. The 521 exclusions comprise 362 assets failing raster, mesh or
rendering checks and 159 held-out icons with file contents identical to a
training icon. The checks reject unstable rasterization, invalid extrusion
geometry and incomplete or misaligned reference renderings.
The retained dataset contains 24,436 glyphs and 2,785 icons, split into
21,904 training, 2,660 validation and 2,657 test assets.
Glyphs are divided by font family in an approximately 8:1:1 ratio, with
206 training families and 25 families in each held-out split. All 77 characters
occur in every split, so the glyph evaluation tests familiar characters in
unseen typefaces.

\begin{figure*}[t]
  \centering
  \begin{subfigure}[b]{0.235\linewidth}
    \centering
    \includegraphics[width=\linewidth]{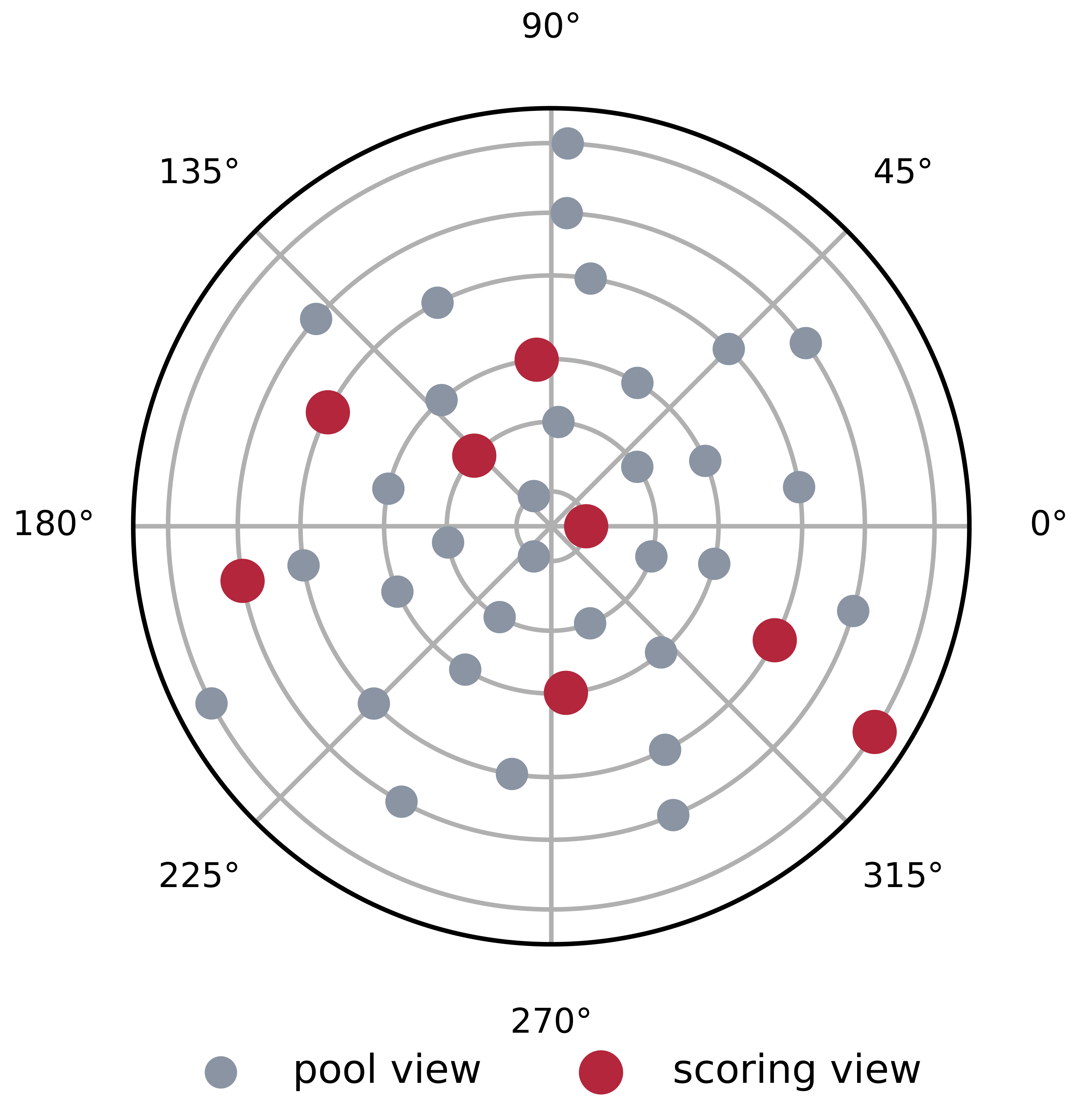}
    \caption{Camera rig.}
    \label{fig:supp-dataset-cameras}
  \end{subfigure}\hfill
  \begin{subfigure}[b]{0.735\linewidth}
    \centering
    \begin{minipage}[b]{0.18\linewidth}
      \centering
      \includegraphics[height=\linewidth]{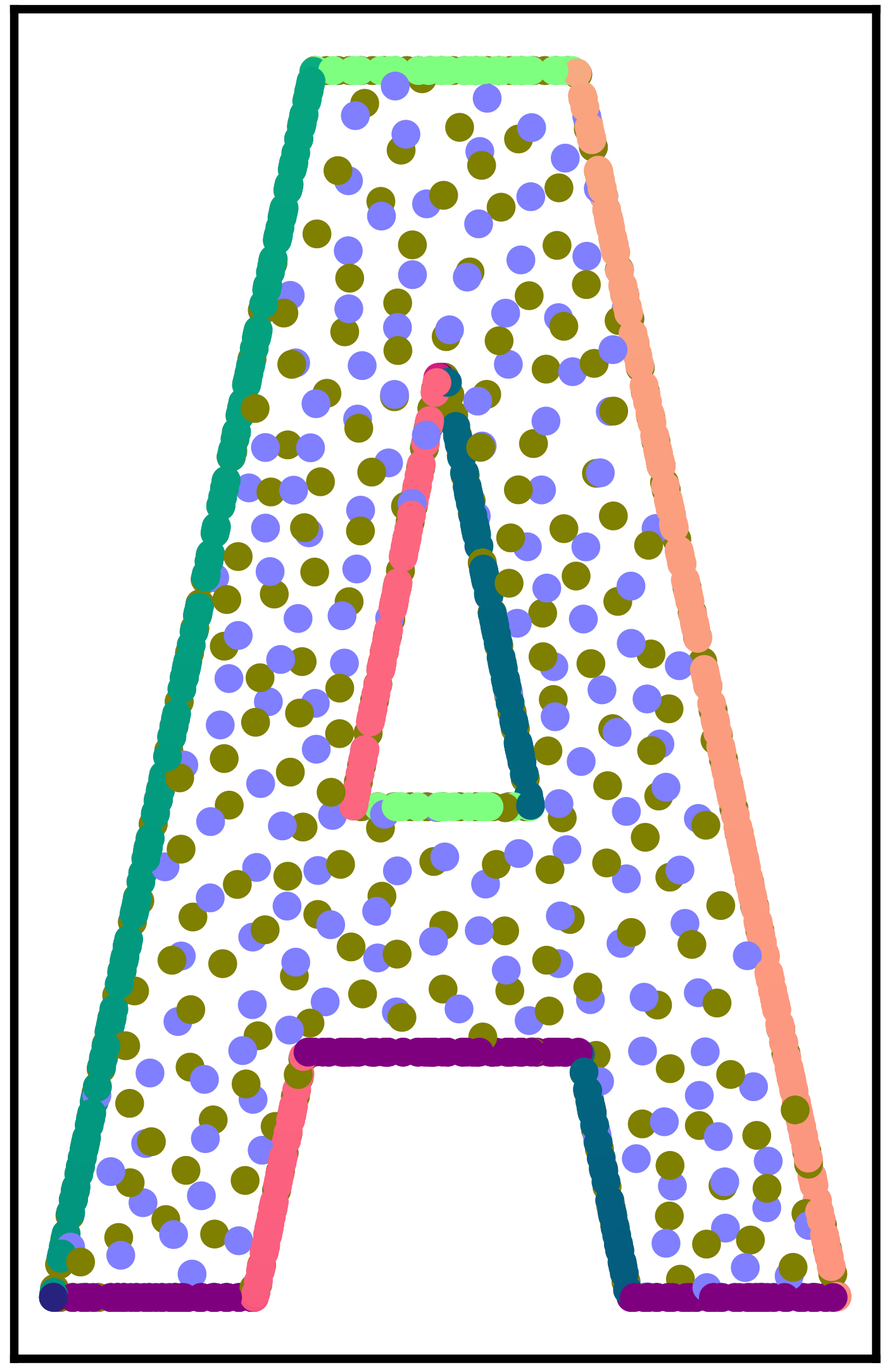}\\[2pt]
      {\scriptsize Anchors ($xy$)}
    \end{minipage}\hfill
    \begin{minipage}[b]{0.18\linewidth}
      \centering
      \includegraphics[height=\linewidth]{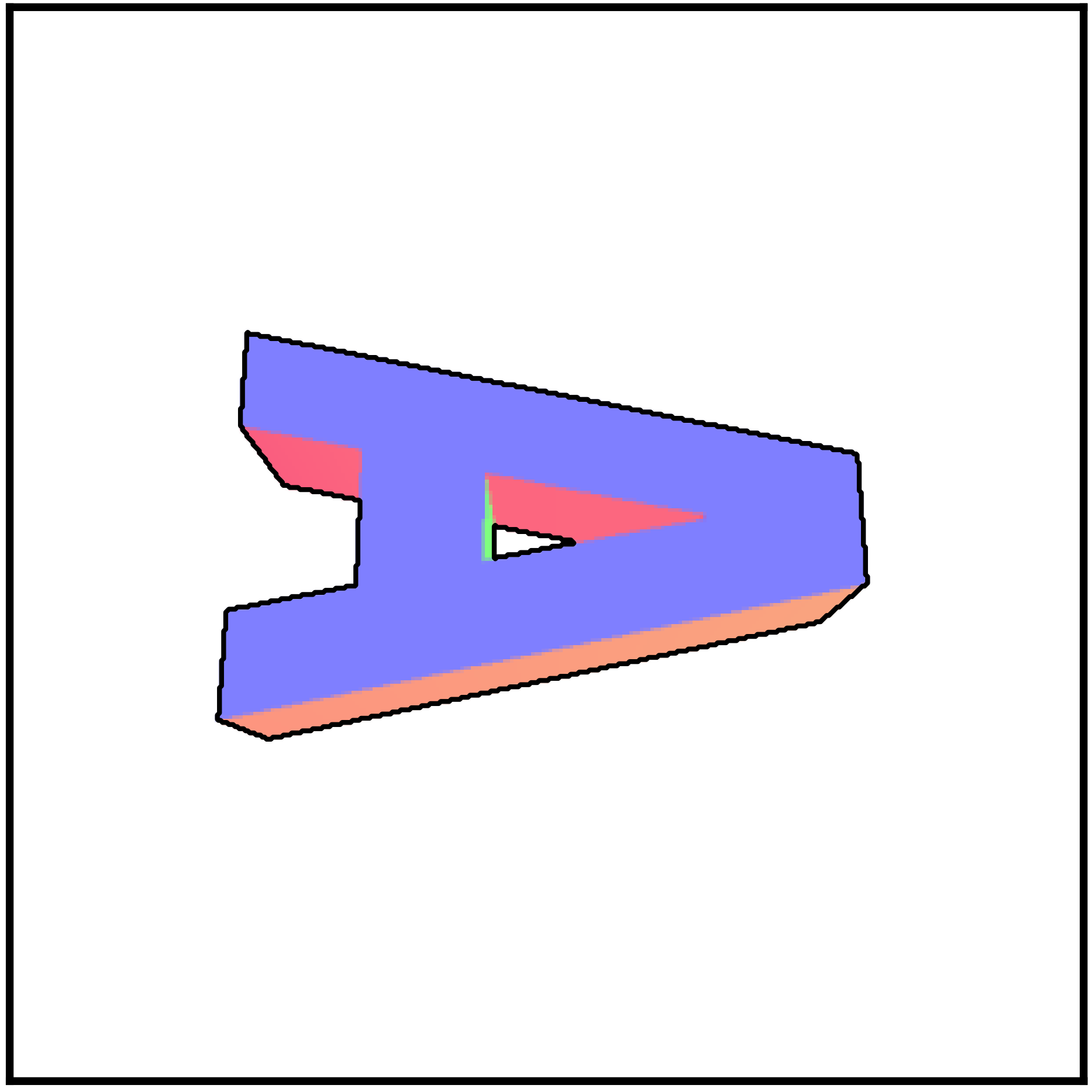}\\[2pt]
      {\scriptsize View 0}
    \end{minipage}\hfill
    \begin{minipage}[b]{0.18\linewidth}
      \centering
      \includegraphics[height=\linewidth]{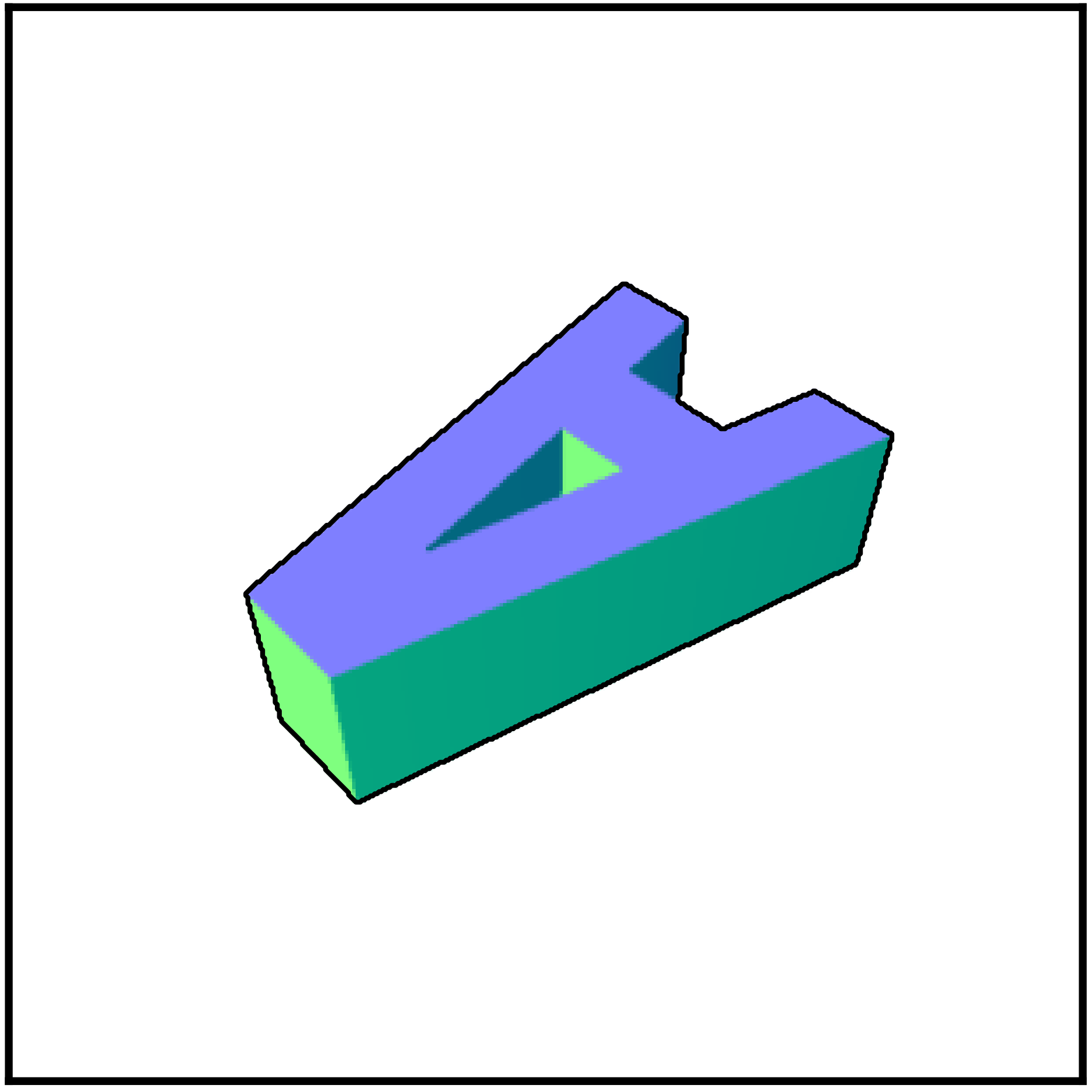}\\[2pt]
      {\scriptsize View 3}
    \end{minipage}\hfill
    \begin{minipage}[b]{0.18\linewidth}
      \centering
      \includegraphics[height=\linewidth]{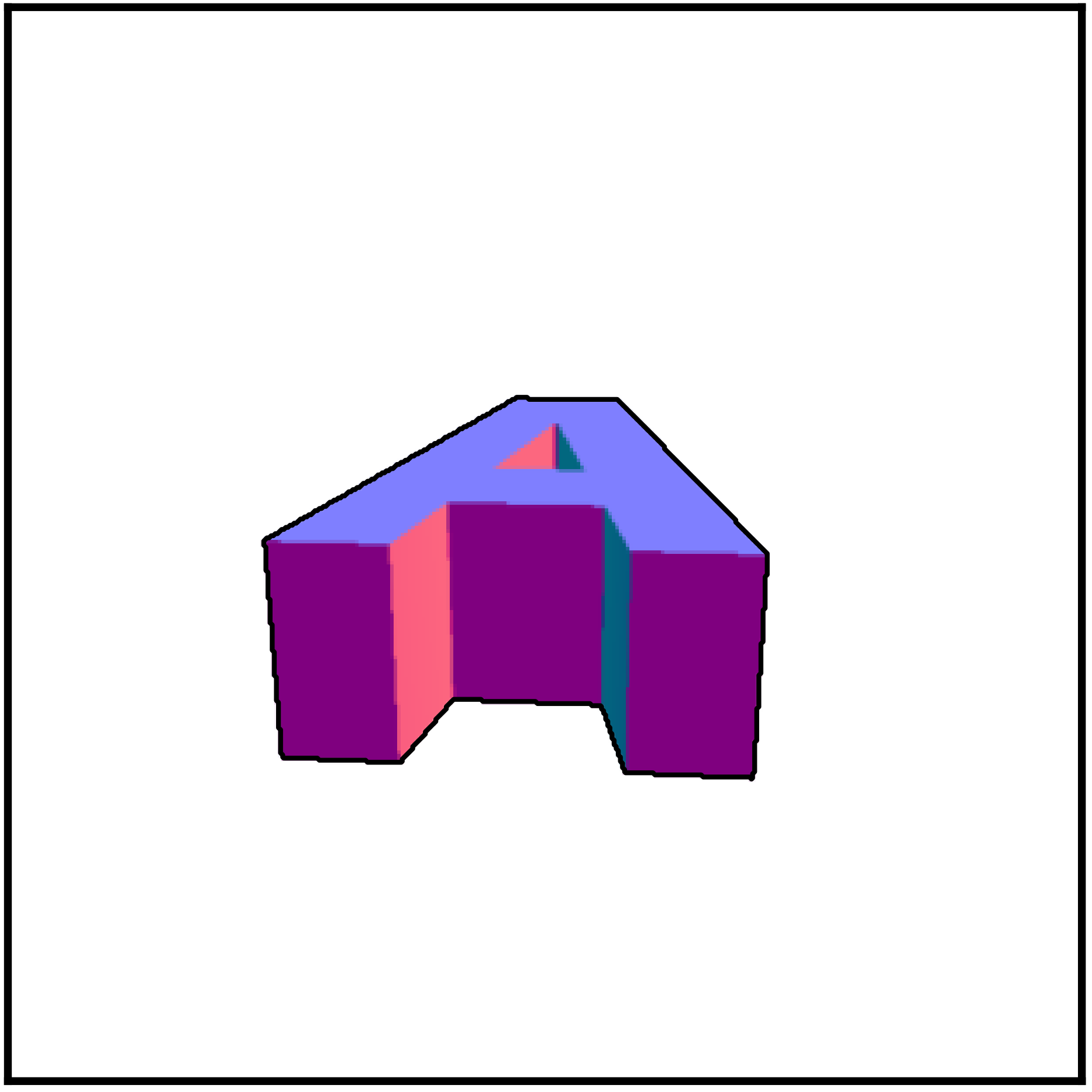}\\[2pt]
      {\scriptsize View 10}
    \end{minipage}\hfill
    \begin{minipage}[b]{0.18\linewidth}
      \centering
      \includegraphics[height=\linewidth]{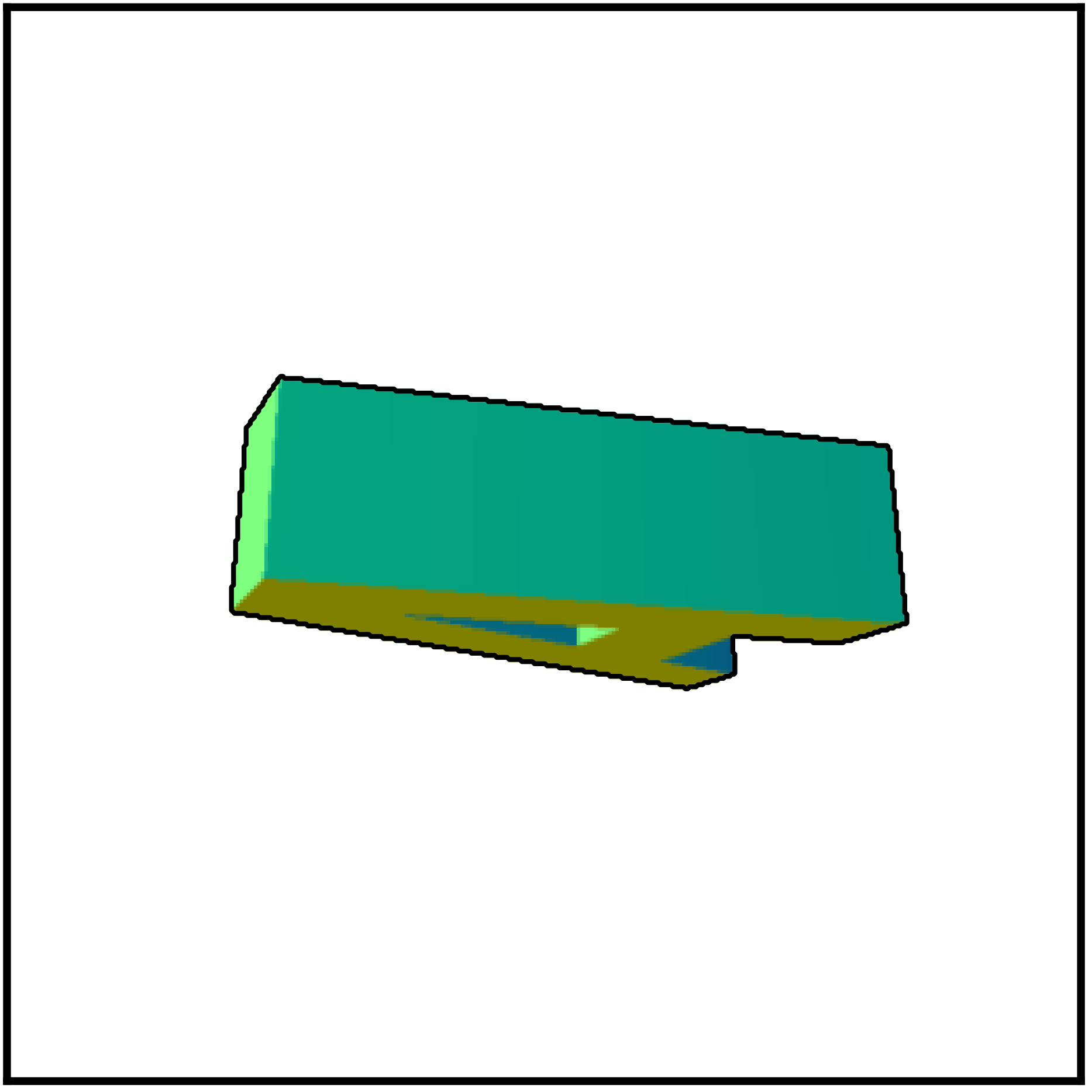}\\[2pt]
      {\scriptsize View 25}
    \end{minipage}
    \caption{Example paired asset.}
    \label{fig:supp-dataset-record}
  \end{subfigure}
  \caption{\textbf{Camera coverage and paired supervision.} (a) The 40-camera rig, with polar angle encoded by radius and azimuth by angle. Red marks the eight held-out (scoring) views, and gray marks the 32 available (pool) views. (b) Projected surface anchors and reference world-space normal maps with silhouette outlines for one glyph at four held-out views.}
  \label{fig:supp-dataset}
\end{figure*}

\noindent\textbf{Cameras and supervision.}
\Cref{fig:supp-dataset} shows the camera coverage and an example of the paired
anchors and multi-view supervision.
All 40 cameras face the origin from radius 1.559 in canonical units, with a
horizontal field of view of $50.81^\circ$. Six rings contain 3, 7, 10, 10, 7 and
3 cameras at polar angles $15^\circ$, $45^\circ$, $72^\circ$, $108^\circ$,
$135^\circ$ and $165^\circ$ from the extrusion axis. Azimuths are equally
spaced within each ring. No silhouette touches the image border in any stored
view. We reserve the first camera in every ring, and the
opposing camera in each ten-camera ring for held-out evaluation, yielding eight views
that cover both caps and the grazing directions along the side walls. They
are excluded from training, fitting, refinement, and the inputs of
image-conditioned baselines.

Blender renders world-space normals and silhouettes at $512^2$ with 64 samples
per pixel. Normal error is evaluated on foreground pixels with valid reference
normals. The flat-region metric used in the supplementary tables covers the
foreground outside the edge band defined in the main paper. Metrics are
averaged over pixels within a view, then over the eight held-out views of each
asset, and finally over assets, so every asset has equal weight.

\section{Baseline adaptations and fitting budgets}
\label{supp:baselines}

\subsection{Baseline inputs and adaptations}

The feed-forward baselines retain their different input requirements while
being retrained on our training split to predict normals and silhouettes.
\Cref{tab:supp-adaptation} records the changes beyond replacing color output
with normals. In comparison, \method{} receives anchor positions alone.
All outputs use the common rasterizer at $512^2$ and are scored on eight
held-out views per asset, averaging first over views and then over assets.

\begin{table*}[!htbp]
\centering\footnotesize
\setlength{\tabcolsep}{4pt}
\caption{\textbf{Feed-forward baseline adaptations.} Inputs retain each converter's geometry or image conditioning, and rendered input views come from the 32 available views. All three converters are retrained with normal targets.}
\label{tab:supp-adaptation}
\begin{tabular}{@{}p{0.17\linewidth}p{0.27\linewidth}p{0.49\linewidth}@{}}
\toprule
Method & Input & Adaptation \\
\midrule
TexGaussian~\cite{xiong2025texgaussian} & Surface points and normals from a regenerated extrusion mesh, organized in the released octree & Normal head replaces color; text, CLIP and material heads removed. Leaf-center placement and identity rotation; our objective and rasterizer; exponential-moving-average weights (rate 0.999) used for scoring. \\
TriplaneGaussian~\cite{zou2024triplanegaussian} & Our anchors and one rendered normal view & Generative point stage bypassed. We implement a training loop for the second-stage regressor, applying its loss family to normals. Geometry and normals are rotated back to world coordinates for scoring. \\
LGM~\cite{tang2024lgm} & Four rendered normal views at ring azimuths & Diffusion front end omitted; four-view regressor retrained for normal output with the released optimizer settings and learning rate $4\times10^{-4}$. \\
\bottomrule
\end{tabular}
\end{table*}

These results characterize task-specific adaptations rather than verified
reproductions of the original methods. LGM's validation error is still
improving at the end of its 30k-step schedule, so the comparison does not
establish a ranking of architectures at convergence.
The regional errors also qualify the overall comparison. TexGaussian has
slightly lower test flat-region error than \method{} ($0.830^\circ$ versus
$0.920^\circ$), despite higher overall and edge-band errors.

\noindent\textbf{Per-asset fitting.}
All three fitting baselines start from the same anchors and optimize against
the 32 available normal and silhouette views.
3DGS~\cite{kerbl20233dgs} initializes Gaussian means at the anchors and adds an
explicit normal attribute, initialized from the analytic face normal and
cloned during densification.
Scaffold-GS~\cite{lu2024scaffoldgs} voxelizes the anchors, replaces its color
head with a normal head of the same shape, and retains native anchor growth
while using our rasterizer, with no color supervision.
Densification in 3DGS and anchor growth in Scaffold-GS stop halfway through
each fitting budget.
Plane-and-decoder fitting optimizes our feature plane and decoder heads from
scratch for each asset, without an encoder. It is a single-plane variant of
TV-GS~\cite{kathariya2026tv}, replacing its two tri-planes with our plane.

We select fitting settings on 16 validation assets by normal error at 32k
steps, using IoU to break ties, and freeze them before test evaluation.
Each method uses its selected settings at all seven test budgets.
Both 3DGS and Scaffold-GS use a densification or growth threshold of
$2\times10^{-3}$ and keep their published learning rates without
scene-scale rescaling. Scaffold-GS uses the released
code's voxelization rule. The selected plane-and-decoder setting uses no
opacity sparsity penalty.
The storage figures quoted in the main comparison are medians over this
64-asset subset, counting half-precision parameter payloads.

\subsection{Fidelity across fitting budgets}

\Cref{tab:supp-fitting-full} extends the main comparison to all seven fitting
budgets on the same 64 test assets, selected at evenly spaced ranks of edge
fraction. The full trajectories show why the largest budget alone does not
summarize fitting quality, since all three methods reach their lowest held-out
normal error before 32k steps. For 3DGS and plane-and-decoder fitting,
silhouette IoU continues to improve as normal error rises.
Scaffold-GS overtakes one-pass conversion in overall normal error at longer
budgets, but \method{} retains lower edge-band error at every fitting budget.

\begin{table*}[!htbp]
\centering\footnotesize
\setlength{\tabcolsep}{5pt}
\caption{\textbf{Per-asset fitting across seven budgets.} Normal, flat-region and edge-band errors and silhouette IoU on eight held-out views of the same 64 test assets. The one-pass reference and fitting results each use one seed. Budgets are optimization steps. \bestkey{Red} and \secondkey{orange} cells mark the best and second-best values across budgets within each fitting method.}
\label{tab:supp-fitting-full}
\begin{tabular}{@{}lrrrrr}
\toprule
Method & Budget & Normal ($^\circ$) $\downarrow$ & Flat ($^\circ$) $\downarrow$ & Edge-band ($^\circ$) $\downarrow$ & IoU $\uparrow$ \\
\midrule
\method{} (ours) & one pass & 1.701 & 0.997 & 8.90 & 0.9840 \\
\midrule
3DGS fitting & 500 & 5.185 & 3.530 & 21.95 & 0.8357 \\
 & 1k & 4.400 & 2.841 & 20.18 & 0.8700 \\
 & 2k & 2.123 & 0.887 & 14.74 & 0.9405 \\
 & 4k & \second 1.931 & \second 0.731 & 14.18 & 0.9577 \\
 & 8k & \best 1.849 & \best 0.701 & \second 13.61 & 0.9681 \\
 & 16k & 1.938 & 0.809 & \best 13.57 & \second 0.9772 \\
 & 32k & 2.099 & 0.956 & 13.97 & \best 0.9811 \\
\midrule
Scaffold-GS fitting & 500 & 4.268 & 2.749 & 19.83 & 0.8863 \\
 & 1k & 3.402 & 1.925 & 18.53 & 0.9144 \\
 & 2k & 2.733 & 1.380 & 16.55 & 0.9400 \\
 & 4k & 1.982 & 0.939 & 12.93 & 0.9680 \\
 & 8k & 1.755 & 0.725 & 11.83 & 0.9794 \\
 & 16k & \best 1.462 & \best 0.555 & \best 10.84 & \best 0.9867 \\
 & 32k & \second 1.514 & \second 0.601 & \second 10.95 & \second 0.9863 \\
\midrule
Plane-and-decoder fitting & 500 & 2.629 & 1.371 & 15.59 & 0.9482 \\
 & 1k & 2.006 & 0.930 & 13.01 & 0.9648 \\
 & 2k & 1.764 & 0.755 & \best 12.09 & 0.9725 \\
 & 4k & \best 1.733 & \second 0.716 & \second 12.14 & 0.9776 \\
 & 8k & \second 1.754 & \best 0.708 & 12.47 & 0.9804 \\
 & 16k & 1.854 & 0.780 & 12.95 & \second 0.9817 \\
 & 32k & 1.897 & 0.789 & 13.35 & \best 0.9828 \\
\bottomrule
\end{tabular}
\end{table*}

\subsection{Gaussians attached to the mesh}
\label{supp:mesh-attached}

Since the extrusion mesh is known, Gaussians can also be attached to it
directly. We test two such methods with their official code:
Mesh2Splat~\cite{scolari2025mesh2splat}, which converts a mesh without
training, and GaMeS~\cite{waczynska2024games}, which fits Gaussians
parameterized on mesh faces for each asset. Both receive the normalized
extrusion meshes used for the TexGaussian inputs and are scored on the same
64 test assets as the fitting comparison.

We run the Mesh2Splat conversion pass without its graphical interface,
supply the mesh's vertex normals, and score the normal it stores with each
Gaussian. Its conversion resolution sets the sampling density. We test ten
resolutions from 16 to 1024 and select the splat scale for each resolution
from five candidates on 16 validation assets. 520 is the default density.
GaMeS fits the 32 available normal views, encoded as color with silhouettes
as alpha, using its own loss and mesh-attached parameterization, with mesh
vertices fixed as in the released code. Its normal is the shortest scale
axis, oriented toward the camera. Because the extrusion meshes contain few
large faces, we also subdivide them until every edge is at most 3\% or 5\%
of the bounding-box diagonal, as the GaMeS documentation recommends for
large faces. Of six configurations, validation selects 3\% subdivision with
three Gaussians per face.
Storage counts position, rotation, scale, opacity and normal in half
precision, 28 bytes per Gaussian, and omits baseline color, in the
baselines' favor.

\Cref{tab:supp-mesh-attached} lists representative settings. With no more
storage than \method{}, Mesh2Splat (resolution 64) and GaMeS on the
unsubdivided meshes have $3.2^\circ$--$13.8^\circ$ higher normal error.
Denser settings reduce the gap, but no Mesh2Splat setting among the ten
resolutions and five splat scales, and no GaMeS configuration at any of the
seven budgets from 500 to 32k steps, matches \method{} in normal error,
edge-band error or IoU. Mesh2Splat has lower flat-region error
from resolution 256 upward, since its splats copy the planar cap normals,
but its edge-band error stays more than twice ours.
On all 2,657 test assets, Mesh2Splat's normal error exceeds ours by
$0.69^\circ$ at resolution 520 and $3.38^\circ$ at resolution 64. The
official code returns no Gaussians for two of these assets; counting them
as failures or patching the read-back changes no comparison.
Mesh2Splat's conversion pass takes 0.13--0.33\,ms per asset on an RTX A6000
at the listed densities, while GaMeS optimizes each asset for up to minutes.

\begin{table*}[!htbp]
\centering\footnotesize
\setlength{\tabcolsep}{5pt}
\caption{\textbf{Gaussians attached to the extrusion mesh.} Median per-asset storage and mean normal, flat-region, and edge-band errors and silhouette IoU on eight held-out views of the 64 test assets. Storage counts half-precision Gaussian parameters without color for the baselines, and our plane and anchors for \method{}. Mesh2Splat settings are conversion resolutions, and GaMeS settings give the mesh, Gaussians per face, and optimization steps. \bestkey{Red} and \secondkey{orange} cells mark the best and second-best values in each error and IoU column.}
\label{tab:supp-mesh-attached}
\begin{tabular}{@{}llrrrrr@{}}
\toprule
Method & Setting & Storage & Normal ($^\circ$) $\downarrow$ & Flat ($^\circ$) $\downarrow$ & Edge-band ($^\circ$) $\downarrow$ & IoU $\uparrow$ \\
\midrule
\method{} (ours) & one pass & 401\,kB & \best 1.701 & 0.997 & \best 8.90 & \best 0.9840 \\
\midrule
Mesh2Splat & 64 & 339\,kB & 4.888 & 2.222 & 32.43 & 0.9372 \\
 & 128 & 1.36\,MB & 3.126 & 1.348 & 21.41 & 0.9617 \\
 & 256 & 5.43\,MB & \second 2.307 & \second 0.602 & 19.73 & 0.9590 \\
 & 520 (default) & 22.5\,MB & 2.371 & \best 0.544 & 20.89 & 0.9535 \\
\midrule
GaMeS & unsubdivided, 2, 32k & 15.2\,kB & 15.472 & 13.605 & 35.12 & 0.9437 \\
 & unsubdivided, 10, 32k & 76.2\,kB & 8.711 & 6.888 & 27.79 & 0.9712 \\
 & 5\% subdivided, 1, 32k & 503\,kB & 3.459 & 2.078 & 17.65 & 0.9720 \\
 & 3\% subdivided, 3, 4k & 4.72\,MB & 2.742 & 1.502 & \second 15.42 & \second 0.9725 \\
 & 3\% subdivided, 3, 32k (selected) & 4.72\,MB & 2.764 & 1.499 & 15.71 & 0.9700 \\
\bottomrule
\end{tabular}
\end{table*}

\section{Network details and optimization protocols}
\label{supp:architecture}

\subsection{Encoder and decoder}
\label{supp:encoder}

The point encoder maps each anchor through
$\mathrm{Linear}(3,96)\to\mathrm{ReLU}\to\mathrm{Linear}(96,96)$ before its
six vector-attention blocks. The same 16-nearest-neighbor graph in anchor
coordinates is used by every block. For relative-position embedding
$\boldsymbol{\delta}_{ij}=\eta(\mathbf a_i-\mathbf a_j)$, a block computes
\begin{equation}
 \boldsymbol{\omega}_{ij}=\operatorname*{softmax}_{j\in\mathcal N(i)}
 \zeta(W_q\mathbf h_i-W_k\mathbf h_j+\boldsymbol{\delta}_{ij}),
\end{equation}
\begin{equation}
 \mathbf h_i\leftarrow\mathrm{LayerNorm}\!\left(\mathbf h_i+
 W_o\sum_{j\in\mathcal N(i)}\boldsymbol{\omega}_{ij}\odot
 (W_v\mathbf h_j+\boldsymbol{\delta}_{ij})\right).
\end{equation}
Here $\eta$ and $\zeta$ are two-layer maps with a ReLU between their
layers and hidden and output width 96, and $\eta$ receives three coordinates.
Attention is normalized over neighbors independently in each channel.
The blocks preserve the anchor count. Two linear maps, $96\to48$ and
$48\to48$, produce the features for plane aggregation. The second is
bias-free and initialized to the identity. Padding used for batching is masked
out of all geometric and feature computations.

After aggregation, the residual plane block applies
$\mathrm{Conv}_{3\times3}(48,64)\to\mathrm{ReLU}\to
\mathrm{Conv}_{3\times3}(64,48)$ with padding one.
Its final convolution is initialized to zero, so the residual block initially
preserves the aggregated plane. The 48 output channels are then split equally
between displacement and attribute features.

\Cref{tab:supp-heads} specifies both the hidden activations and the output
transformations of the shared decoder. Write
$\mathbf x_i^{\mathrm{disp}}=[\mathbf f_i^{\mathrm{disp}};\mathbf a_i]$ and
$\mathbf x_i^{\mathrm{att}}=[\mathbf f_i^{\mathrm{att}};\mathbf a_i]$.
Only opacity and covariance append the camera-to-anchor direction
$\mathbf d_i(v)$. Each head has hidden width 96 and produces attributes for
six candidate Gaussians per anchor, except for the scale head, which produces
two three-dimensional scales shared by those candidates.

\begin{table*}[t]
\centering\small
\caption{\textbf{Decoder heads and output transformations.} All heads have one hidden layer of width 96. Transformations apply after the final linear layer. The covariance head produces three scale factors and four quaternion components per candidate Gaussian.}
\label{tab:supp-heads}
\begin{tabular}{@{}llcll@{}}
\toprule
Head & Input & Output width & Hidden activation & Output transformation \\
\midrule
Offset & $\mathbf x_i^{\mathrm{disp}}$ & 18 & Sine & Linear \\
Anchor scales & $\mathbf x_i^{\mathrm{disp}}$ & 6 & ReLU & Linear \\
Opacity & $[\mathbf x_i^{\mathrm{att}};\mathbf d_i(v)]$ & 6 & ReLU & Sigmoid \\
Covariance & $[\mathbf x_i^{\mathrm{att}};\mathbf d_i(v)]$ & 42 & ReLU & Sigmoid (scale); unit normalization (rotation) \\
Normal & $\mathbf x_i^{\mathrm{att}}$ & 18 & ReLU & Unit normalization \\
\bottomrule
\end{tabular}
\end{table*}

The sine activation in the offset head uses unit frequency. The scale head's two outputs
supply $\mathbf s_i^{\mathrm{off}}$ and $\mathbf s_i^{\mathrm{base}}$,
and the covariance scale factors multiply the latter. The opacity head starts
with zero final-layer weights and bias $\operatorname{logit}(0.03)$.
The encoder and decoder contain 466,624 and 22,746 parameters, respectively.

\noindent\textbf{Decoder width in the latent controls.}
The tri-plane controls concatenate the features queried from all three planes.
With eight channels per plane half, each decoder half therefore receives
24 feature values, as in the single plane. With 24 channels per plane half,
it receives 72. This wider input accompanies the lower-resolution $37^2$
tri-plane control, so the comparison changes resolution, per-plane channel
count and decoder width together and does not isolate spatial resolution.

\noindent\textbf{Default training.}
\label{supp:training}
We train the default encoder and decoder jointly for 30k steps with Adam
(default moments) and no weight decay. Both parameter groups start at a learning rate of
$2\times10^{-3}$, with a half-cosine schedule that decays to $10^{-4}$
without warm-up.

\subsection{Refinement protocol}
\label{supp:refinement}

We optimize the predicted plane after the residual block, keeping the encoder
and decoder fixed. Adam (default moments) uses a constant step size of $3.2\times10^{-3}$ and
four available views per update, with the objective and weights in the main
paper. Both choices are selected on 16 validation assets before test
refinement. The per-anchor-feature control instead optimizes its own
predicted features with a separately selected step size of $10^{-2}$.
These features improve and then deteriorate over the budgets in the same way
as the plane, so a short refinement budget serves both representations.

The elapsed-time budget covers optimization updates, excluding encoding and
scoring, and is measured after ten untimed warm-up steps with device
synchronization. The 1, 10 and 60\,s endpoints belong to the same trajectory.
The 10\,s refinement measurement used an RTX~A4000 and averages
385 updates, and is distinct from the RTX~A6000 forward-pass timing in the
main comparison. Endpoints are evaluated after storing the optimized features
in half precision. The stored plane is the refined state, since rerunning the
encoder would restore the original prediction.

\subsection{Plane prediction and decoder sharing}
\label{supp:study}

The bounded study in the main paper compares predicted planes with a shared
decoder (A), fitted planes with a shared decoder (B), and fitted planes with
separate decoders (C). Its 32 training assets comprise 28 glyphs from distinct
families and four icons, stratified across geometric difficulty. A disjoint
32-asset subset is used to select step sizes. Plane dimensions, decoder
architecture, objective, and view access are shared across all three settings.
Within each seed, the decoders share their initialization, and B and C use
the same random initial plane for each asset.

At each training cycle, we sample 16 distinct views per asset from the
32 available views. The draw is refreshed each cycle and shared across
A, B, and C.
A and B minimize the mean loss over all 32 assets in one accumulated optimizer
step. In B, the 32 planes are independent trainable tensors, and each receives
one update without an encoder evaluation. C trains 32 separate plane--decoder
pairs serially, updating each once per cycle. The 16k-cycle budget thus gives
every plane and decoder 16k updates in all settings, although C executes
32 separate optimizer steps per cycle. All results use the final checkpoint
and eight held-out views of these training assets, over three seeds.

The selected initial step sizes are $2\times10^{-3}$ for A and
$6.3\times10^{-3}$ for B and C, each decayed to 5\% by a cosine schedule.
On an RTX~A5000, median cycle times are 2.24, 1.66 and 1.81\,s for A, B and C.
A additionally evaluates and backpropagates through the encoder, whereas B
and C optimize their planes directly. Its higher training cost in the
wall-clock plot therefore accompanies a different constraint on the latent,
not extra view exposure. This study measures held-out-view fidelity after
training on the same assets. The unseen-family experiment tests generalization
to new assets separately.

\section{Additional architecture and loss controls}
\label{supp:ablations}

\begin{figure*}[!tp]
  \centering
  \includegraphics[width=\linewidth]{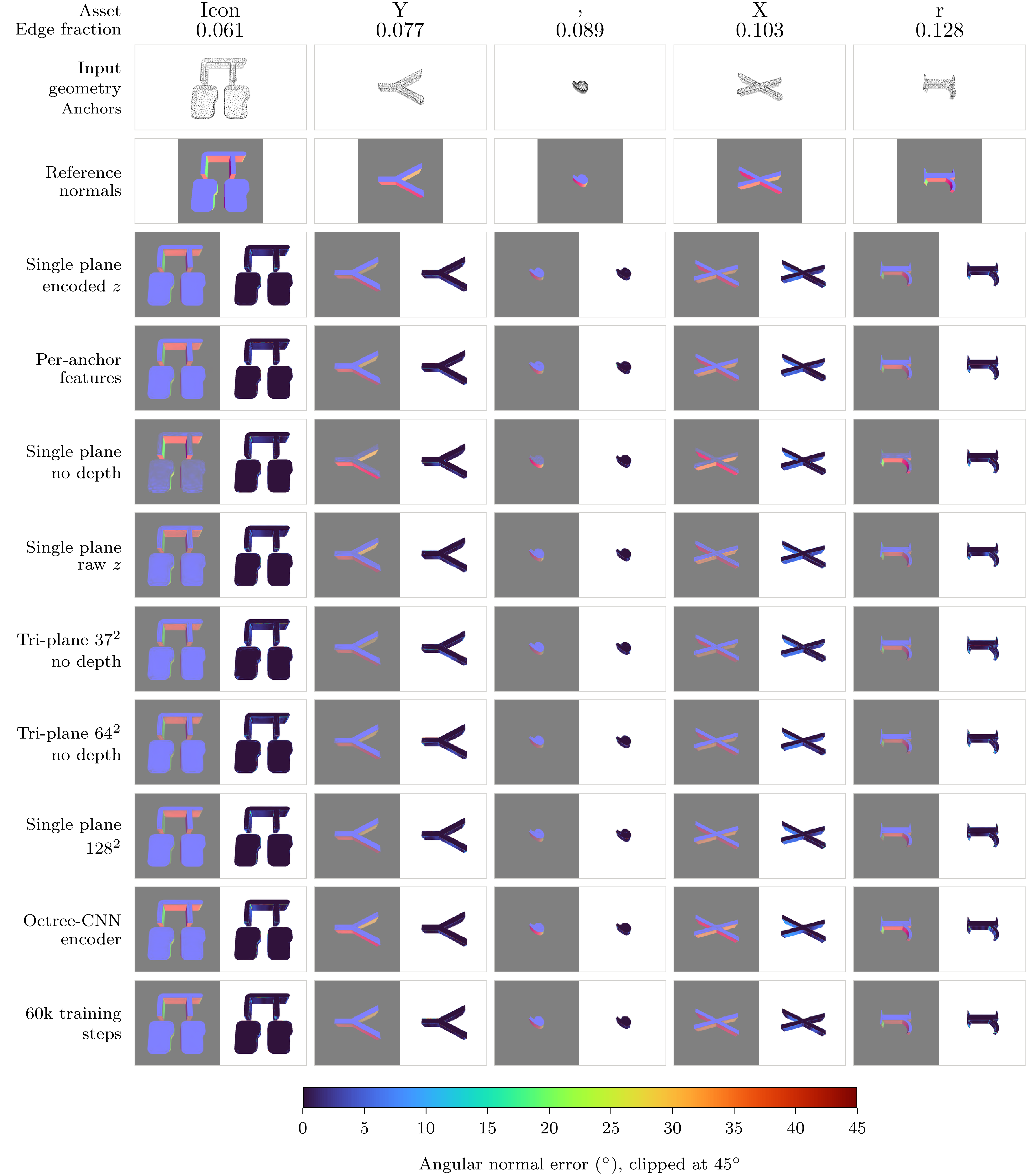}
  \caption{\textbf{Architecture controls on validation examples.} Five assets ordered by edge fraction, shown at one held-out view using one training seed. The first two rows show projected anchors and reference normals, and each subsequent cell pairs rendered normals (left) with angular error (right), clipped at $45^\circ$. The tri-plane rows shown here omit depth conditioning, distinct from the depth-conditioned tri-plane controls in \cref{tab:architecture}.}
  \label{fig:supp-architecture}
\end{figure*}

\Cref{tab:supp-roster} reports the resolution and training-length controls
mentioned in the main paper. These extend the $64^2$ plane with the sinusoidal
depth conditioning $\gamma(z)=[\sin\pi z,\cos\pi z,\sin 2\pi z,\cos 2\pi z]$. The raw-depth default and other latent layouts are
compared in the main paper. All results use 2,660 validation assets and
held-out views. The finer grid costs four times the latent storage and
mainly reduces edge-band error. Longer training mainly reduces flat-region
error. Both give small changes in overall normal error relative to the
reference's $0.132^\circ$ standard deviation across seeds.

\begin{table}[!htbp]
\centering\footnotesize\setlength{\tabcolsep}{3pt}
\caption{\textbf{Resolution and training length.} Means over three seeds on the validation split. The reference uses a $64^2$ plane, $\gamma(z)$, and 30k steps. \bestkey{Red} and \secondkey{orange} cells mark the best and second-best distinct values per metric.}
\label{tab:supp-roster}
\begin{tabular}{@{}lrrrr}
\toprule
Variant & \shortstack{Normal\\($^\circ$) $\downarrow$} & \shortstack{Flat\\($^\circ$) $\downarrow$} & \shortstack{Edge-band\\($^\circ$) $\downarrow$} & IoU $\uparrow$ \\
\midrule
Reference & 1.675 & 0.930 & 8.92 & \second 0.9842 \\
$128^2$ plane & \second 1.631 & \second 0.928 & \best 8.49 & 0.9841 \\
60k steps & \best 1.630 & \best 0.885 & \second 8.88 & \best 0.9853 \\
\bottomrule
\end{tabular}
\end{table}

\Cref{tab:supp-losscontrols} reports the encoder and loss controls against
their common single-seed reference. Replacing the point transformer with an
Octree-CNN raises normal error under the same training recipe, despite
similar backbone sizes (0.40\,M parameters). The encoder change also alters
spatial discretization, and its learning rate is not retuned. Removing the SSIM term ($\mathcal{L}_{\mathrm{ssim}}$) increases normal error,
primarily through the flat regions' contribution to the mean. Removing the scale regularizer instead improves
both fidelity metrics in this seed, so these controls do not demonstrate a
fidelity benefit from that regularizer. We retain it because it stabilizes
training and discourages degenerately large Gaussians.

\begin{table}[!htbp]
\centering\footnotesize\setlength{\tabcolsep}{3pt}
\caption{\textbf{Encoder and loss controls.} Single-seed results on the same validation split. The reference uses the point transformer and full objective, and each control changes the named component. \bestkey{Red} and \secondkey{orange} cells mark the best and second-best values per metric.}
\label{tab:supp-losscontrols}
\begin{tabular}{@{}lrrrr}
\toprule
Variant & \shortstack{Normal\\($^\circ$) $\downarrow$} & \shortstack{Flat\\($^\circ$) $\downarrow$} & \shortstack{Edge-band\\($^\circ$) $\downarrow$} & IoU $\uparrow$ \\
\midrule
Reference & \second 1.712 & \second 0.960 & \second 9.03 & 0.9845 \\
Octree-CNN & 2.005 & 1.149 & 10.27 & 0.9838 \\
No SSIM term & 1.908 & 1.147 & 9.31 & \second 0.9857 \\
No scale regularizer & \best 1.579 & \best 0.869 & \best 8.48 & \best 0.9864 \\
\bottomrule
\end{tabular}
\end{table}

\begin{samepage}
\noindent\textbf{Qualitative comparison.}
\Cref{fig:supp-architecture} locates the errors on five validation assets
spanning the edge-fraction range. The plane, per-anchor, and tri-plane rows
compare feature layouts and depth inputs, and the last three rows show the
resolution, encoder, and training-length controls. Raw depth is the default,
while encoded depth is the reference for the resolution and training-length
comparisons. Each prediction is paired with its angular
error map, making cap and side-wall errors directly comparable. Removing
depth conditioning visibly disrupts the first icon's silhouette, while the other
controls retain its broad shape, with errors concentrated near surface creases.

\end{samepage}

\section{Family-level evaluation}
\label{supp:family}

The seen--unseen comparison weights font families equally, so typefaces with
more files or retained glyphs do not dominate the result. The
geometry-adjusted comparison additionally accounts for each family's mean
edge fraction. For a family $f$, let $e_f$ be this fraction, $u_f$ indicate
an unseen family, and $y_f$ denote its mean normal error or IoU. We fit
\begin{equation}
 y_f=\beta_0+\beta_e e_f+\beta_u u_f+\epsilon_f
\end{equation}
by unweighted least squares over the 206 seen and 25 unseen families, and
report $\beta_u$ as the adjusted gap. Edge fraction is computed from reference
images alone. The adjusted normal-error gap has magnitude $0.04^\circ$.

For the near-duplicate check, the front silhouette is taken from the held-out
camera at $15^\circ$ to the extrusion axis. We crop it to its bounding box,
resize its longer side to 256 pixels while preserving aspect ratio, and
center it on a $256^2$ canvas. Each test glyph is compared with all training
glyphs of the same character, and its maximum silhouette IoU determines
whether it is a near-duplicate. Excluding matches above $0.98$ removes 476
of the 2,445 test glyphs (19.5\%). Four families have no remaining glyphs, leaving 1,969 glyphs in
21 unseen families. The 206-family reference is unchanged. This exclusion
applies only to the sensitivity analysis.
The resulting normal-error gap has magnitude $0.02^\circ$, so removing the closest
silhouette matches preserves the small seen--unseen difference.
The oblique reference view includes side-wall pixels, which can raise IoU
between distinct designs and make the exclusion conservative relative to a
strictly axial silhouette test.

\section{Coding and delivery}
\label{supp:delivery}

Delivery stores the feature plane and anchors so that a client can reconstruct the asset with the shared Gaussian decoder.
\Cref{fig:delivery} in the main paper plots the HEVC rate--distortion results, and this section gives the coding settings and the values at each operating point.
We adopt feature-image coding from TV-GS~\cite{kathariya2026tv}. Compatibility here concerns standard image and video codecs, not TV-GS bitstreams or decoder weights.
Reported total sizes are the measured lengths of the plane and anchor files, including headers and quantizer side information, with $1\,\mathrm{kB}=1000$ bytes.
They exclude the decoder, stored once and shared by all assets, and the material assigned at use time.
The encoder is needed to predict a plane, but not to render a delivered asset.
The coding experiments use one training seed for each model.

\subsection{Quantization and reconstruction}
For the default and refined planes, a feature value $y$ is mapped to a ten-bit code using a fixed clip range $\tau=2.5$:
\begin{equation}
\begin{aligned}
  q(y)&=\operatorname{round}\!\left(\frac{1023}{2\tau}
  \big[\operatorname{clip}(y,-\tau,\tau)+\tau\big]\right),\\
  \hat y&=\frac{2\tau}{1023}\hat q-\tau .
\end{aligned}
\end{equation}
Here $\hat q$ is the code recovered by the codec, and rounding uses ties to even.
The clip range is fixed using 16 validation assets before evaluating the 64 test assets used for the fitting comparison.

The 48 channels form $64\times64$ tiles in eight columns and six rows, with the 24 displacement channels followed by the 24 attribute channels.
They fill the luma plane of a $512\times384$ frame with ten-bit 4:2:0 sampling and constant chroma values of 512.
We code each asset as one frame with HM-18.0, the HEVC reference encoder, using its all-intra main-10 configuration.
Per-frame quantization offsets are zero, and the in-loop, sample-adaptive-offset and temporal filters are disabled.
The eight QPs are $7,12,\ldots,42$.
The matching decoder recovers the frame, which is untiled and inverse-quantized to reconstruct the feature plane.

Anchor positions are quantized to a lattice of step $2^{-14}$, merging coincident cells.
A hybrid octree codes occupancy after Morton ordering, and LZMA2 at preset 9 compresses the payload when this reduces its size.
The reconstructed positions and plane feed the shared Gaussian heads for the requested camera.
Anchor quantization changes normal error by at most $0.004^\circ$ on an individual asset.
The anchor file averages 5.8\,kB and is included at every operating point, accounting for about half the total at 11.5\,kB.
\Cref{tab:supp-hevc} lists the mean total size and fidelity at each QP for the three variants in \cref{fig:delivery}.

\begin{table*}[!htbp]
\centering\footnotesize
\caption{\textbf{HEVC operating points on 64 test assets.} Mean total size of the coded plane and anchor files, held-out normal error and silhouette IoU at each QP. The uncoded row evaluates each variant's plane before quantization. Refined planes are the exported 10\,s states.}
\label{tab:supp-hevc}
\begin{tabular}{@{}lrrrrrrrrr@{}}
\toprule
& \multicolumn{3}{c}{Default} & \multicolumn{3}{c}{Refined (10\,s)} & \multicolumn{3}{c}{Rate-aware} \\
\cmidrule(lr){2-4}\cmidrule(lr){5-7}\cmidrule(l){8-10}
QP & Total (kB) & Normal ($^\circ$)$\downarrow$ & IoU$\uparrow$ & Total (kB) & Normal ($^\circ$)$\downarrow$ & IoU$\uparrow$ & Total (kB) & Normal ($^\circ$)$\downarrow$ & IoU$\uparrow$ \\
\midrule
Uncoded & -- & 1.701 & 0.9840 & -- & 1.313 & 0.9825 & -- & 1.635 & 0.9854 \\
\midrule
7  & 52.0 & 1.702 & 0.9840 & 54.7 & 1.316 & 0.9824 & 67.8 & 1.635 & 0.9854 \\
12 & 42.6 & 1.706 & 0.9839 & 45.3 & 1.321 & 0.9824 & 57.8 & 1.636 & 0.9854 \\
17 & 32.7 & 1.719 & 0.9838 & 35.3 & 1.342 & 0.9823 & 47.5 & 1.636 & 0.9854 \\
22 & 23.9 & 1.755 & 0.9835 & 26.3 & 1.409 & 0.9822 & 38.3 & 1.642 & 0.9853 \\
27 & 16.6 & 1.916 & 0.9822 & 18.0 & 1.609 & 0.9811 & 29.2 & 1.658 & 0.9853 \\
32 & 11.5 & 2.653 & 0.9713 & 12.0 & 2.397 & 0.9721 & 20.6 & 1.722 & 0.9852 \\
37 & 8.5 & 5.511 & 0.9281 & 8.6 & 5.166 & 0.9317 & 13.7 & 1.941 & 0.9844 \\
42 & 6.8 & 10.913 & 0.8876 & 6.9 & 10.547 & 0.8874 & 9.4 & 2.773 & 0.9809 \\
\bottomrule
\end{tabular}
\end{table*}

The same quantized frame can be stored with a lossless still-image codec. We test PNG, JPEG~XL~\cite{alakuijala2019jpegxl}, JPEG~2000~\cite{skodras2001jpeg2000} and AVIF, which uses AV1 intra coding~\cite{han2021av1}.
\Cref{tab:supp-stillimage} shows that all four preserve the ten-bit codes exactly and thus preserve their rendered fidelity.
Lossless coding preserves the quantized representation, not the original floating-point values.

\begin{table}[!htbp]
\centering\footnotesize
\caption{\textbf{Lossless image delivery on 64 test assets.} Each codec exactly preserves the ten-bit plane codes, giving identical fidelity. Total storage includes the unchanged anchor file. \bestkey{Red} and \secondkey{orange} cells mark the smallest and second-smallest totals. PNG stores the codes in a 16-bit container, and JPEG~2000 uses its reversible mode.}
\label{tab:supp-stillimage}
\begin{tabular}{@{}lrrr@{}}
\toprule
Codec & Total (kB) $\downarrow$ & Normal ($^\circ$) $\downarrow$ & IoU $\uparrow$ \\
\midrule
PNG & 108.6 & 1.701 & 0.9840 \\
JPEG~XL & \best 72.4 & 1.701 & 0.9840 \\
JPEG~2000 & 95.0 & 1.701 & 0.9840 \\
AVIF & \second 92.3 & 1.701 & 0.9840 \\
\bottomrule
\end{tabular}
\end{table}

\subsection{Refinement at equal delivered size}
Delivery of a refined asset requires exporting its optimized plane, since rerunning the encoder would recover the initial prediction.
We code the exported 10\,s refinement states and measure coding distortion against those same states before coding.
Although refinement leaves the plane dimensions unchanged, it increases coded plane storage by up to 13\% at equal QP.

To compare at equal size, we use each asset's default operating points as targets and linearly interpolate refined normal error in log total file size between adjacent points of that asset's refined coding curve.
We average the paired error differences only over assets whose curves cover the target size, without extrapolation.
The sizes below are means of the per-asset default target sizes.
All 64 assets overlap at the default operating points from 52.0 down to 8.5\,kB.
Refinement lowers mean normal error at every default point from 52.0 down to 16.6\,kB, by margins shrinking from $0.385^\circ$ to $0.146^\circ$.
At 11.5\,kB the interpolated difference is $+0.091^\circ$, with a 95\% paired bootstrap interval of $[0.001, 0.175]^\circ$ over assets, an estimate sensitive to linear interpolation across the steep part of each refined curve.
At 8.5\,kB the estimated difference of $+0.092^\circ$ is inconclusive, since its interval of $[-0.045, 0.224]^\circ$ includes zero.
At the smallest default point, 6.8\,kB, only two assets overlap, so it does not support a subset-wide comparison.
The refinement benefit at equal size therefore holds down to 16.6\,kB.

\subsection{Optional rate-aware training}
\label{supp:rate-aware}
The default model applies quantization and coding only after training.
The optional \emph{rate-aware} variant follows TV-GS~\cite{kathariya2026tv}, adding uniform noise in $[-\tfrac12,\tfrac12]$ to plane values during training and a Gaussian-bin rate term to \cref{eq:loss} with weight $\lambda_{\mathrm{rate}}=5\times10^{-4}$.
This term is the negative log probability mass of a unit-width bin around each noisy value, using one learned mean and standard deviation per plane half across the dataset.
Only results labeled \emph{rate-aware} use this additional objective.
For half $h\in\{\mathrm{disp},\mathrm{att}\}$, let $y_{hj}$ index its $N_{\mathrm{p}}=24\times64^2$ entries and let $\tilde y_{hj}=y_{hj}+\epsilon_{hj}$, with $\epsilon_{hj}\sim\mathcal{U}(-\tfrac12,\tfrac12)$.
With $\Phi$ the standard Gaussian cumulative distribution function,
\begin{equation}
\begin{aligned}
 p_{hj}={}&\Phi\!\left(\frac{\tilde y_{hj}+\tfrac12-\mu_h}{\sigma_h}\right)
       -\Phi\!\left(\frac{\tilde y_{hj}-\tfrac12-\mu_h}{\sigma_h}\right),\\
 \mathcal{L}_{\mathrm{rate}}={}&-\frac{1}{2N_{\mathrm{p}}}
 \sum_{h\in\{\mathrm{disp},\mathrm{att}\}}\sum_{j=1}^{N_{\mathrm{p}}}\log p_{hj} .
\end{aligned}
\end{equation}
Each half has one learned pair $(\mu_h,\sigma_h)$, initialized at $(0,0.1)$.
The same noisy planes enter the decoder, coupling this rate proxy to rendered fidelity.
The variant is jointly trained from scratch for 30k steps, with the rate term active throughout at weight $5\times10^{-4}$. The remaining training settings follow \cref{supp:training}.
Noise and the rate term are disabled at evaluation, and the learned density parameters are not exported.

The rate-aware planes require a different quantizer. Validation selects per-channel ranges rather than the default fixed clipping, adding 384 bytes of side information per asset to the reported totals.
At equal total size, the measured normal error is $0.041$--$0.066^\circ$ lower than the default from 16.6\,kB upwards, with all 64 assets covered, a gap already present before coding (\cref{tab:supp-hevc}).
At 11.5\,kB the reduction is $0.240^\circ$, but only 58 assets overlap. At 8.5\,kB only four overlap.

\section{Shading and appearance control}
\label{supp:appearance}

Appearance assignment uses the deferred metallic-roughness shader adopted from PBR-TV-GS~\cite{lee2026pbrtvgs}.
After Gaussian rasterization, each foreground pixel is shaded using its normalized world-space normal, viewing direction, and the asset's homogeneous albedo, roughness, and metallicity.
The Cook--Torrance model combines a diffuse lobe with GGX specular reflection, Schlick Fresnel, and height-correlated Smith visibility.
Directional lights supply a direction and radiance, while point-light directions are evaluated per pixel.
Spatially constant ambient illumination and material emission can also contribute.
Shading is independent of network training, so materials and lighting can be changed without converting the geometry again.
The resulting nonnegative radiance is transformed to sRGB and composited with the background using the rendered alpha.

Environment illumination uses either irradiance represented by nine spherical-harmonic coefficients per color channel or a prefiltered equirectangular stack.
For the stack, diffuse illumination samples the blurriest level along the normal and is weighted by the nonmetallic albedo and one minus the Fresnel term.
Glossy illumination samples along the reflected view direction at a level selected by roughness and is weighted by Fresnel.
This approximation omits the second factor of the split-sum approximation and evaluates Fresnel with the normal--view cosine.

The shader evaluates pixels independently, without visibility queries between the asset, lights, and surrounding scene.
Smith visibility represents microfacet masking, not scene occlusion.
Consequently, this appearance control does not model self-shadowing, cast or received shadows, interreflection, or asset--scene light transport.
Materials are homogeneous within each asset.
Depth ordering during insertion provides geometric occlusion but does not add these illumination interactions.

\subsection{Supplementary videos}
\label{supp:videos}
Two videos, available on the project page, animate the scene of \cref{fig:teaser} for 10\,s at 30 frames per second and $1280\times720$ resolution, keeping its asset placements, materials, and lights, and starting from its first view.
In \texttt{coffee-fixed.mp4} the camera stays at this view, and in \texttt{coffee-wide-left.mp4} it moves to a viewpoint on the left of the scene and back.
The letters hop in turn, the heart pulses, $\mathrm{F}_1$ and $\mathrm{F}_2$ spin in opposite directions, and the star slides behind the glass and back.
The animation changes only instance transforms and light positions, so the converted assets are not predicted again.
The background plays the first 64 frames of the Coffee Martini sequence~\cite{li2022n3dv}, reconstructed with TV-GS~\cite{kathariya2026tv}, forward and then backward.

\end{document}